\documentclass[preprint,12pt]{elsarticle}

\usepackage{amssymb}
\usepackage{amsmath}
\usepackage{booktabs}  
\usepackage{multirow}
\usepackage{amssymb}  
\usepackage{amsmath}
\usepackage{algorithm}        
\usepackage{algpseudocode}    

\usepackage{graphicx} 
\usepackage{hyperref}
\usepackage{algorithm}
\usepackage{algpseudocode}
\usepackage{bm}

\usepackage{booktabs}
\usepackage{graphicx}
\usepackage{amsmath}
\usepackage[table]{xcolor} 
\usepackage{soul} 

\usepackage{subfigure}

\journal{arxiv}

\begin{document}

\begin{frontmatter}

\title{DFed-SST: Building Semantic- and Structure-aware Topologies for Decentralized Federated Graph Learning}


\author{Lianshuai Guo$^a$} 
\author{Zhongzheng Yuan$^a$} 
\author{Xunkai Li$^b$} 
\author{Yinlin Zhu$^c$} 
\author{Meixia Qu$^a$\textsuperscript{*}} 
\author{Wenyu Wang$^a$\textsuperscript{*}} 
\affiliation{organization={Shandong University, School of Mechanical, Electrical and Information Engineering},
            city={Weihai},
            postcode={264209}, 
            country={China}}
\affiliation{organization={Beijing Institute of Technology, School of Computer Science and Technology},
            city={Beijing},
            postcode={100081}, 
            country={China}}
\affiliation{organization={Sun Yat-sen University, School of Computer Science and Engineering},
            city={Guangzhou},
            postcode={510275}, 
            country={China}}

\begin{abstract}
Decentralized Federated Learning (DFL) has emerged as a robust distributed paradigm that circumvents the single-point-of-failure and communication bottleneck risks of centralized architectures. However, a significant challenge arises as existing DFL optimization strategies, primarily designed for tasks such as computer vision, fail to address the unique topological information inherent in the local subgraph. Notably, while Federated Graph Learning (FGL) is tailored for graph data, it is predominantly implemented in a centralized server-client model, failing to leverage the benefits of decentralization.To bridge this gap, we propose DFed-SST, a decentralized federated graph learning framework with adaptive communication.  The core of our method is a dual-topology adaptive communication mechanism that leverages the unique topological features of each client's local subgraph to dynamically construct and optimize the inter-client communication topology. This allows our framework to guide model aggregation efficiently in the face of heterogeneity. Extensive experiments on eight real-world datasets consistently demonstrate the superiority of DFed-SST, achieving  3.26\% improvement in average accuracy over baseline methods.
\end{abstract}










\end{frontmatter}
\section{Introduction}
\label{introduction}
Graph Neural Networks (GNNs) have emerged as a powerful paradigm for processing and analyzing graph-structured data, achieving remarkable success in a myriad of domains such as financial analysis \citep{li2025financial}, molecular structure analysis \citep{ekstrom2023accelerating}, and recommendation systems \citep{wu2022graph}. However, in many real-world scenarios, graph data is not centrally stored but is instead distributed across different institutions or user devices, forming data silos. Examples include transaction networks across different banks or patient case graphs among various hospitals. Centralizing this data for model training raises significant privacy and security concerns.

To address this challenge, Federated Learning (FL) \citep{mcmahan2017communication} was introduced as a viable solution for collaborative modeling while preserving data privacy. This paradigm trains models locally and exchanges only model parameters, such as gradients or weights. The fusion of FL with GNNs, termed Federated Graph Learning (FGL), has consequently become a research hotspot in the graph learning field.

Currently, the majority of mainstream FGL methods adopt a star-shaped topology reliant on a central server \citep{beltran2023decentralized}. In this architecture, the central server distributes the global model, collects and aggregates model updates from all clients, and then disseminates the newly aggregated model. Although this centralized paradigm is conceptually straightforward, it suffers from several inherent deficiencies. First, the central server bears an immense communication load, easily becoming a system-wide bottleneck. Second, it presents a single point of failure; the entire federated system would be paralyzed if the central server fails. Finally, the unconditional trust required in the central server poses potential privacy risks in certain scenarios. To overcome these limitations, Decentralized Federated Learning (DFL) has garnered increasing attention as a more robust and scalable alternative \citep{lalitha2018fully,li2025centralized}, wherein clients communicate and aggregate models directly via a peer-to-peer (P2P) network, completely eliminating the need for a central server. The versatility of this paradigm is evidenced by its successful application in diverse domains. For instance, in the Internet of Things (IoT) domain, \cite{geng2024privacy} proposed a federated learning framework that integrates blockchain and secure multi-party computation, aiming to realize a decentralized and privacy-preserving network to address security challenges. BrainTorrent\citep{roy2019braintorrent} developed a serverless, dynamic peer-to-peer environment that allows all participants to interact directly and collaboratively train brain segmentation models.For aviation equipment fault diagnosis, \cite{MAO2025102876} proposed a federated graph network that enables collaborative diagnosis across heterogeneous data islands by employing a self-perception layer to fuse multi-sensor data and a prototype-based matching mechanism to align client feature spaces. Furthermore, for Unmanned Aerial Vehicle (UAV) networks, \cite{qu2022decentralized} proposed a decentralized federated learning architecture that enables a UAV swarm to perform collaborative learning without a central entity, thereby mitigating the latency issues inherent in traditional cloud-centric models.

The convergence of this decentralized paradigm and FGL establishes our field of study: Decentralized Federated Graph Learning (DFGL). The primary objective of DFGL is to collaboratively train GNN models in a serverless environment. However, in pursuit of this goal, existing research confronts two major limitations.

\begin{figure}[t]
 \centering
 \includegraphics[width=0.95\textwidth]{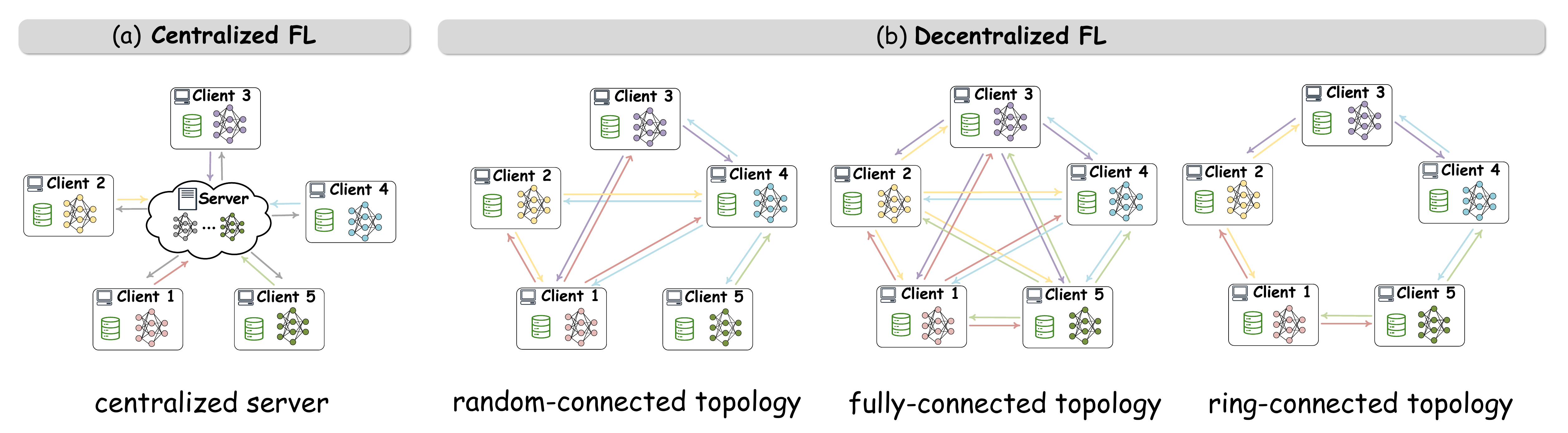}%
 \caption{Illustration of Centralized Federated Learning and various communication topologies of Decentralized Federated Learning.}
 \label{various_topology}
 \vspace{-10pt}
\end{figure}

\begin{figure*}[t]

\rmfamily
\centering
\subfigure[Label Distributions\label{fig:es_sub1}]{
\includegraphics[width=0.45\textwidth]{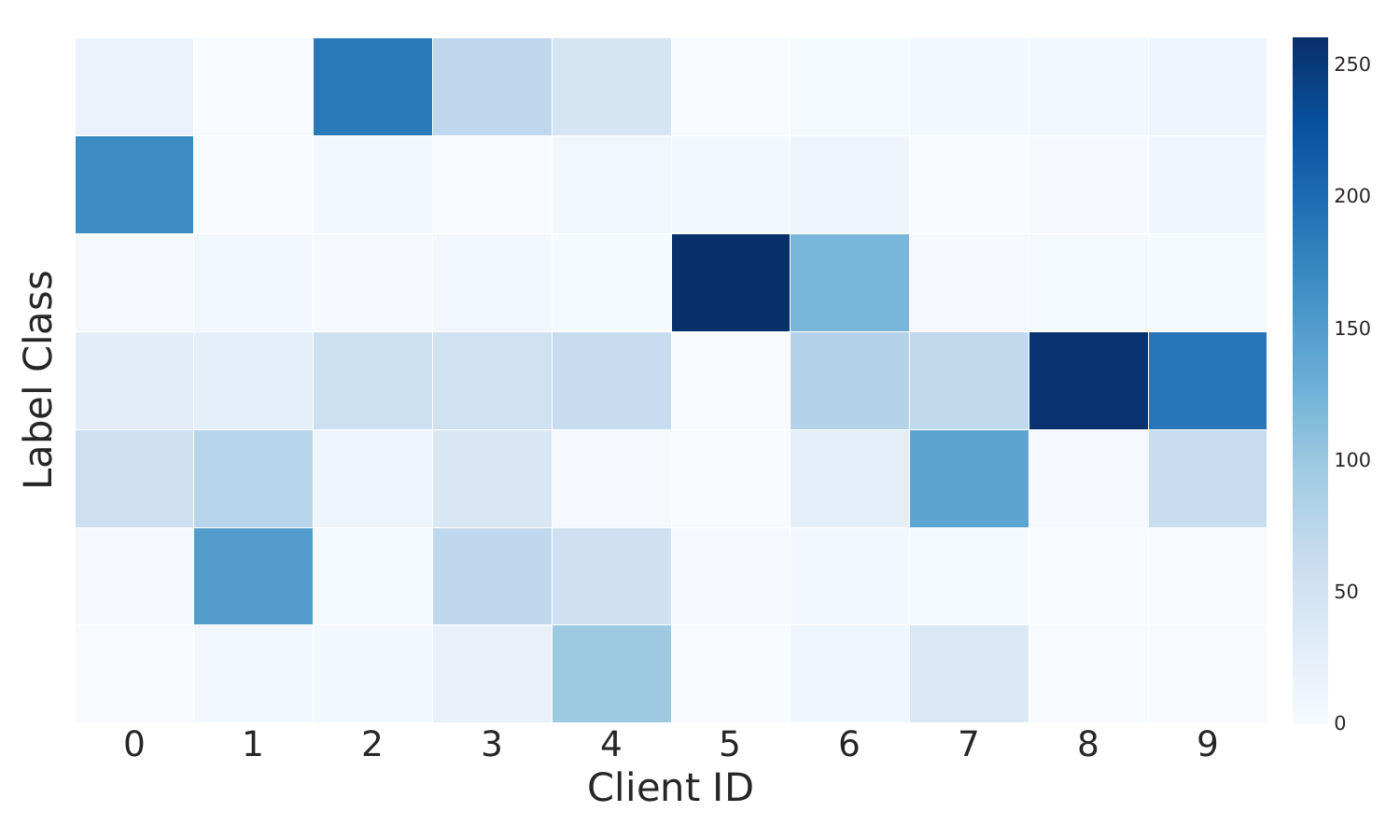} 
}
\subfigure[Homophily Distributions\label{fig:es_sub2}]{
\includegraphics[width=0.45\textwidth]{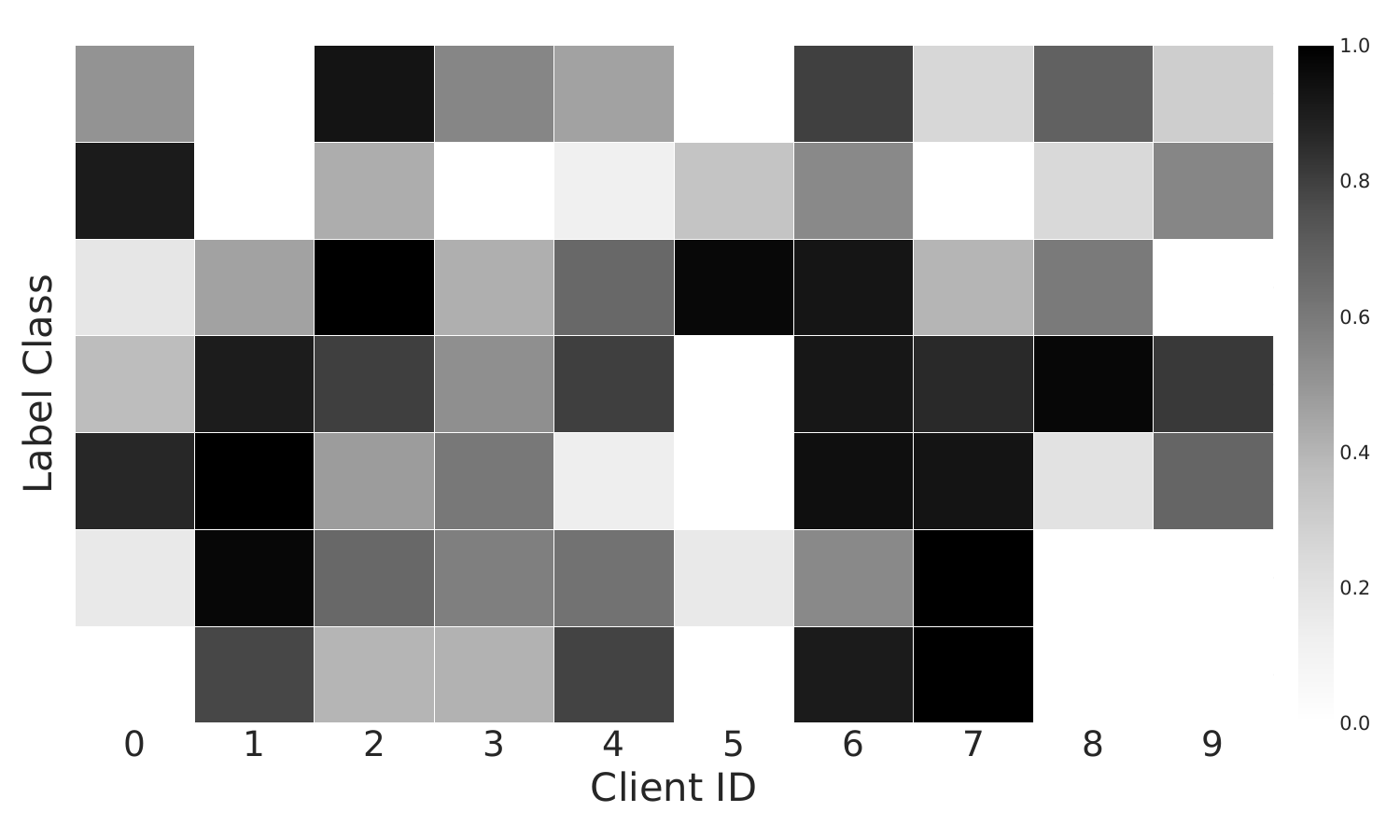} 
}
\hspace{-2mm}
\subfigure[Structural Metrics\label{fig:es_sub3}]{
\includegraphics[width=0.45\textwidth]{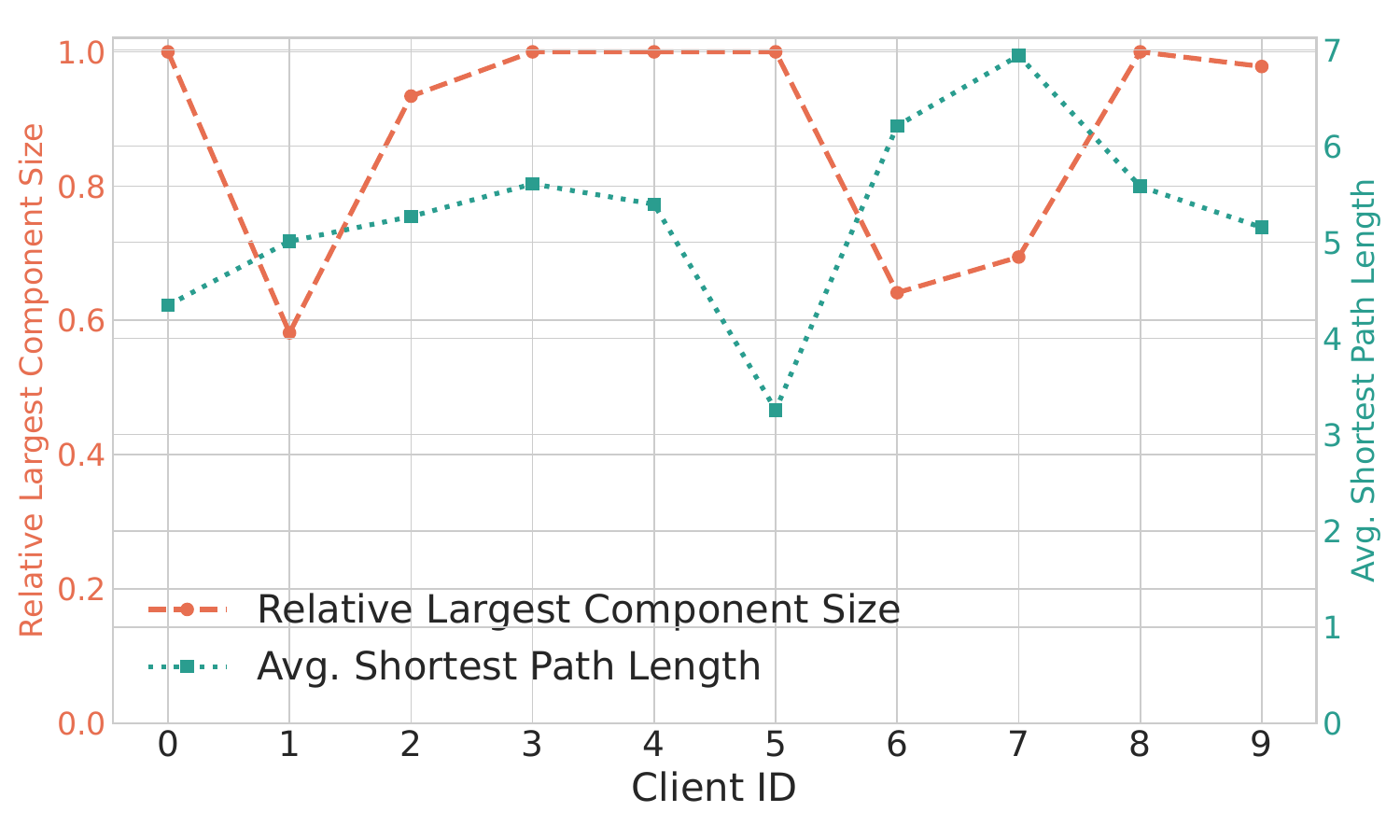} 
}
\subfigure[Performance Comparison\label{fig:es_sub4}]{
\includegraphics[width=0.45\textwidth]{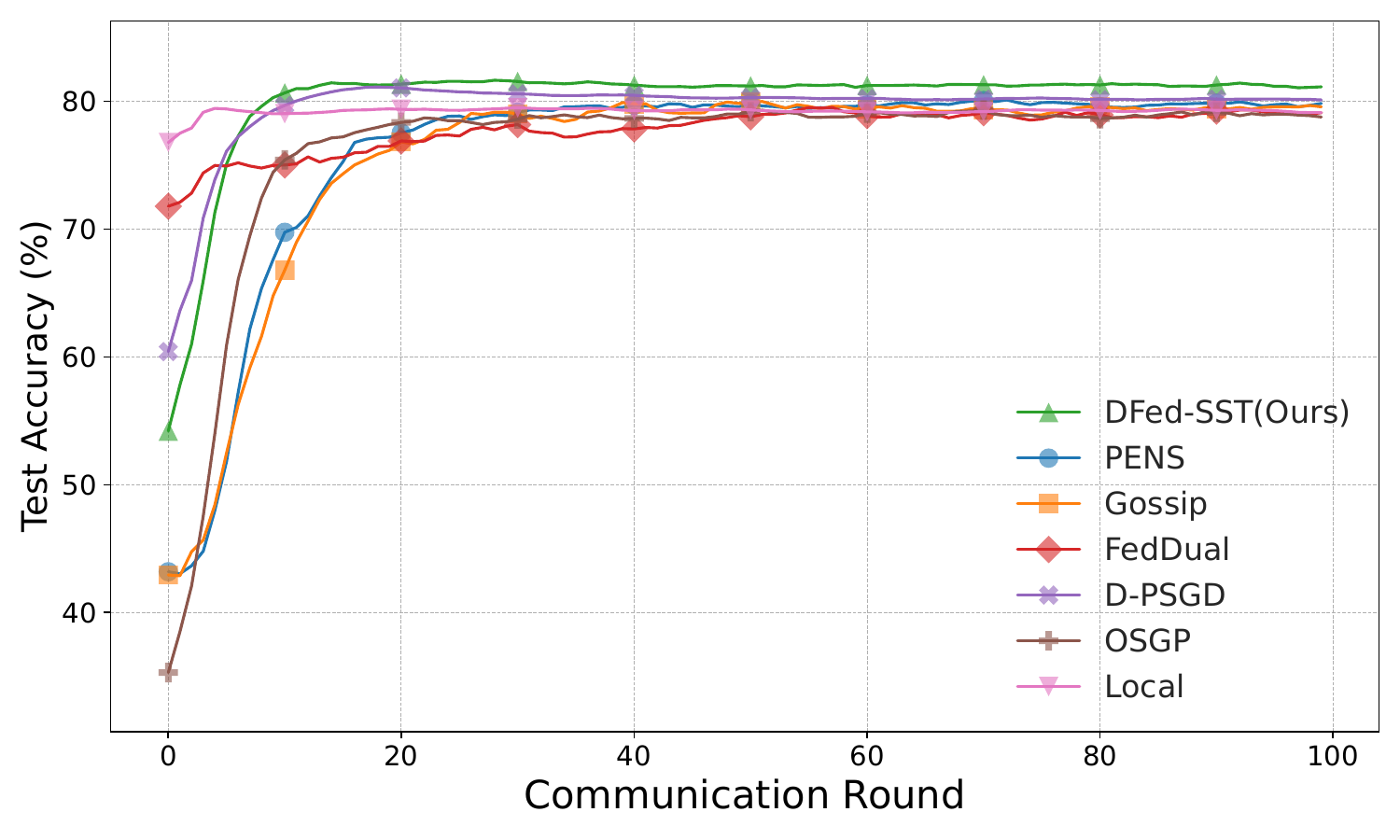} 
}\hspace{-2mm}

\caption{Empirical analysis on the Cora dataset. (a) Heatmap of the number of nodes per class for each client. (b) Heatmap of the intra-class homophily ratio per class for each client. (c) Comparison of structural metrics across clients. (d) Average test accuracy of existing methods over communication rounds, where `Local' represents the scenario with no communication between clients.}
\label{empirical_result}
\end{figure*}

\textit{\textbf{Limitation 1:}} Existing research primarily focuses on designing inter-client communication and aggregation strategies, which are largely determined by the underlying communication topology. The communication topologies in current DFGL research, as illustrated in Fig.~\ref{various_topology}, are typically undirected and symmetric. Structurally, these can be categorized as either static or stochastic dynamic. Regardless of whether they employ a fixed static communication topology (e.g., ring or fully-connected)  \citep{zhu2022topology,pei2021decentralized} or a randomly evolving one \citep{chen2022feddual,hu2019decentralized}, they share a common fundamental limitation: the communication topology is not constructed with client heterogeneity in mind.

\textit{\textbf{Limitation 2:}} Like FGL, DFGL must confront the challenge of subgraph heterogeneity, a point intuitively demonstrated by our empirical analysis on the Cora dataset (see Fig.~\ref{empirical_result}). The label distribution heatmap in Fig.~\ref{fig:es_sub1} reveals significant statistical heterogeneity (Non-IID) among clients. More critically, the node homophily ~\citep{pei2020geom} heatmap for each label class (Fig.~\ref{fig:es_sub2}), combined with the structural metrics ~\citep{xie2021federated} for each client's subgraph such as average shortest path length and maximum component size (Fig.~\ref{fig:es_sub3}), jointly confirms the existence of the profound and unique topological heterogeneity inherent in DFGL. This dual heterogeneity poses a severe challenge to existing optimization strategies. As shown in Fig.~\ref{fig:es_sub4}, existing decentralized methods fail to achieve competitive results when compared to a `local' setting with no inter-client communication. Even with the application of classic decentralized methods, performance remains far from ideal, revealing the limitations of current approaches in handling such complex heterogeneity.

To precisely address this challenge, we propose DFed-SST, a novel Decentralized Federated Graph Learning framework guided by an adaptive topology. Its key insight lies in integrating the dual semantic-structural characteristics of clients into every stage of communication topology construction. This is accomplished through two specifically designed, cooperative modules: the Weighted Label Spatial Dispersion (WLSD) module, which quantifies a client's information complexity to adaptively allocate its number of connections, and the Class-wise Semantic Embedding (CSE) module, which generates a unique semantic-structural fingerprint for selecting the most compatible communication partners.

In summary, the contributions of this paper are three-fold:

\begin{itemize}
 \item \textbf{New Perspective and Problem Formulation}. We introduce a new, data-driven paradigm for DFGL that shifts the focus to dynamic topology construction. Unlike traditional methods, we explicitly define the core optimization objective as ``how to construct a dynamic communication topology that can perceive and adapt to client heterogeneity."

\item \textbf{Novel Framework and Method}. We propose DFed-SST, a novel and adaptive DFGL framework. This framework comprises two core modules: a novel metric, WLSD, to quantify client complexity for adaptive degree determination, and a structure-aware semantic similarity matching mechanism, CSE, for accurately selecting the most relevant communication partners. These two modules work in concert to dynamically generate the communication topology.

\item \textbf{State-of-the-Art Performance}. We conducted comprehensive experiments on eight real-world graph datasets characterized by high heterogeneity. The results confirm the effectiveness and superiority of our proposed method, which significantly outperforms various existing baselines in terms of average model accuracy (a \textbf{3.26\%} improvement).
\end{itemize}

\section{ Preliminary and Related Work}
\label{Preliminary and Related Work}

\subsection{Problem Formulation} 
The core focus of this paper is Decentralized Federated Graph Learning (DFGL). We consider a decentralized system comprising $N$ clients, denoted as $\mathcal{C} = \{C_1, C_2,\cdots, C_N\}$, where no central server exists. Each client $i \in \{1,2,\cdots, N\}$ possesses a local, private graph $\mathcal{G}_i = (\mathcal{V}_i, \mathcal{E}_i, \mathbf{X}_i)$, where $\mathcal{V}_i$ is the set of nodes, $\mathcal{E}_i$ is the set of edges, and $\mathbf{X}_i$ is the node feature matrix. Each graph $\mathcal{G}_i$ is also associated with a set of node labels $\mathbf{Y}_i$.
The system's objective is to collaboratively train a set of high-performance Graph Neural Network (GNN) models, $\{\mathbf{w}_1, \mathbf{w}_2, \cdots, \mathbf{w}_N\}$, through inter-client collaboration without the direct sharing of local graph data, where $\mathbf{w}_i$ represents the local model parameters for client $i$.

In this decentralized setting, inter-client communication is governed by a communication graph $\mathcal{G}_c$ \citep{beltran2023decentralized}. Participants independently and iteratively execute a three-step process: Step 1: each client trains its local model using its private data; Step 2: guided by the communication graph $\mathcal{G}_c$, each client exchanges model parameters with its neighbors; and Step 3: each client aggregates the received parameters to update its local model, preparing for the next round.

Unlike many works that assume a static and undirected communication graph, we consider a more general and realistic scenario by modeling the network as a time-varying, directed graph $\mathcal{G}_c(t) = (\mathcal{C}, \mathcal{E}_c(t))$. Here, $\mathcal{C} = \{C_1,\cdots, C_N\}$ is the set of clients, and $\mathcal{E}_c(t) \subseteq \mathcal{C} \times \mathcal{C}$ is the set of directed communication links active at round $t$. A directed link $(j, i) \in \mathcal{E}_c(t)$ indicates that client $j$ can transmit information to client $i$ during that round, which does not necessarily imply the existence of a reverse link $(i, j)$. Accordingly, we define the set of in-neighbors for client $i$ at round $t$ as $\mathcal{N}_i^{\text{in}}(t) = \{j \mid (j, i) \in \mathcal{E}_c(t)\}$ and its set of out-neighbors as $\mathcal{N}_i^{\text{out}}(t) = \{j \mid (i, j) \in \mathcal{E}_c(t)\}$.

In summary, the core problem this paper addresses is: given the aforementioned DFGL setting, how to devise an efficient strategy to dynamically construct the communication topology $\mathcal{G}_c(t)$ in each round, so as to most effectively leverage the heterogeneous information among clients and enhance final model performance. Our proposed framework, DFed-SST, directly tackles this challenge of dynamic topology construction by offering a novel solution driven by the data characteristics of individual clients.

\subsection{Federated Graph Learning} 
Federated Graph Learning (FGL) applies the principles of Federated Learning to graph-structured data. A core FGL scenario involves subgraph-based federated learning, where multiple clients, each holding a local subgraph of a larger graph, collaboratively perform node- or edge-level tasks. This paradigm primarily confronts two core challenges: subgraph heterogeneity and the loss of inter-client edge information.

To address subgraph heterogeneity, researchers have proposed various topology-aware aggregation strategies. For instance, FedGTA \citep{li2023fedgta} utilizes mixed moments to quantify topological similarity, which in turn guides the aggregation process. Similarly, Fed-PUB \citep{baek2023personalized} employs subgraph embeddings for similarity-based aggregation. FedTAD \citep{zhu2024fedtad} takes this a step further by decoupling node and topological variations to deeply analyze heterogeneity, designing a topology-aware knowledge distillation strategy. Furthermore, to achieve finer-grained personalization, AdaFGL \citep{li2024adafgl} adopts knowledge distillation for client-specific training processes.

To mitigate the issue of missing edge information arising from distributed graph storage, another line of work is dedicated to reconstructing or completing missing structural information. Representative work, such as FedSage+ \citep{zhang2021subgraph}, employs a mechanism for generating missing neighbors to complete the receptive fields of nodes. Similarly, FedGNN \citep{wu2021fedgnn} endeavors to reconstruct potential missing edges between clients by exchanging information, in an attempt to realize a secure federated graph recommendation model.

Despite these significant contributions, the aforementioned methods are predominantly designed for centralized federated frameworks. This centralized architecture not only introduces communication bottlenecks and single-point-of-failure risks but also renders them ill-suited for the fully decentralized network environment that is the focus of our work.

\subsection{Decentralized Federated Learning} 

Decentralized Federated Learning (DFL) was proposed to overcome the communication bottlenecks and single-point-of-failure risks inherent in the centralized paradigm. In the DFL setting, the system no longer relies on a central server; instead, clients, organized in a peer-to-peer network, communicate directly with their neighbors to collaboratively train their models. Many foundational DFL algorithms can be categorized under Decentralized Stochastic Gradient Descent (D-SGD) \citep{assran2019stochastic,sun2022decentralized}. For ease of theoretical analysis, these works often assume that clients perform synchronous or asynchronous model updates on a pre-defined static topology (e.g., ring or grid). To enhance information flow and system robustness, another major line of work is based on gossip-based learning algorithms. For example, Gossip Averaging \citep{hu2019decentralized} and its variants allow clients to perform pairwise model averaging with only one or a few randomly selected neighbors in each round. Information, much like a rumor, eventually diffuses throughout the network via these stochastic, repeated pairwise interactions. FedDual \citep{chen2022feddual} proposes a pairwise gossip mechanism to implement DFL in large-scale decentralized networks while ensuring data privacy. PENS \citep{onoszko2021decentralized} allows clients to actively probe and select peers with the most similar data distributions as neighbors by evaluating their model's loss on each other's data. The DFedPGP \citep{liu2024decentralized} framework tackles heterogeneity in DFL through directed collaboration and partial gradient pushing, achieving better personalization and faster convergence.

These methods have made significant progress in mitigating statistical heterogeneity. However, they perceive heterogeneity through a singular dimension, focusing almost exclusively on statistical variations caused by factors such as non-uniform label distributions. This narrow focus exposes profound limitations when these methods are applied to the multifaceted challenges of graph learning.

\section{Methodology}
\label{Methodology}

\begin{figure}[t]
 \centering
 \includegraphics[width=1.0\textwidth]{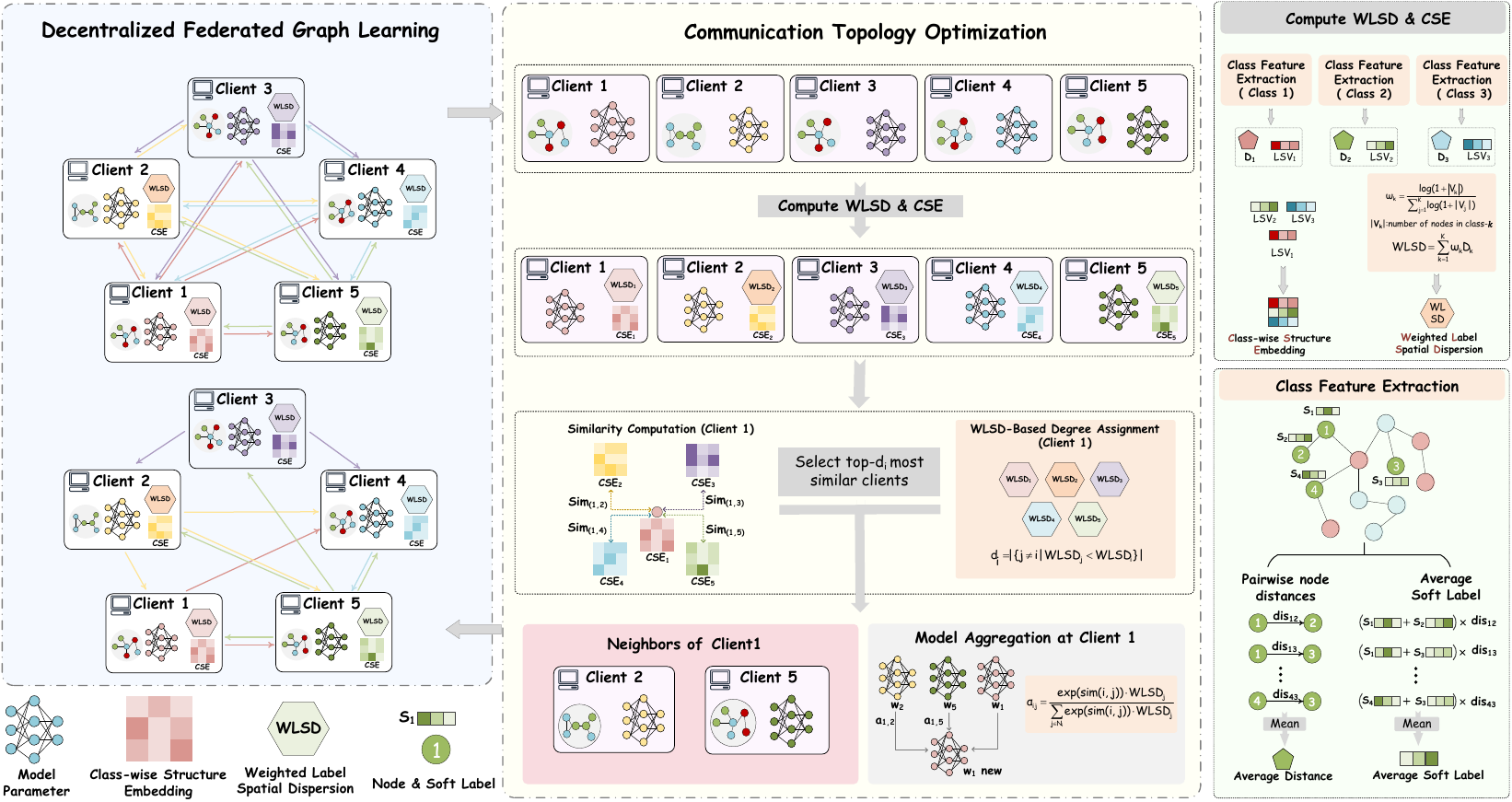}%
 \caption{An overview of our proposed DFed-SST framework.}
 \label{framework}
 \vspace{-10pt}
\end{figure}

To address the heterogeneity challenge in Decentralized Federated Graph Learning, we propose a novel adaptive topology-guided averaging framework, named \textbf{DFed-SST}. The core idea of this framework is to move beyond fixed or random communication patterns, instead constructing a communication topology that dynamically adapts to the dual semantic and structural characteristics of each client. As illustrated in Fig.~\ref{framework}, our method enables clients in a classic decentralized scenario to optimize their communication topology. This is achieved through three core modules: (1) the proposed Weighted Label Spatial Dispersion (WLSD) metric quantifies the information complexity of each client to adaptively determine its communication requirements; (2) Class-wise Semantic Embedding (CSE) generates a unique ``fingerprint'' for each client to enable precise matching with the most relevant partners; and (3) a weighted model aggregation scheme, guided by both similarity and complexity, is performed on the dynamically constructed topology. The entire mechanism operates autonomously without a central coordinator, thereby exhibiting excellent scalability and adaptability to heterogeneous environments. In this section, we elaborate on the details of each core design.

\subsection{Weighted Label Spatial Dispersion}

 \noindent \textbf{Motivation}. As confirmed by our empirical validation in the Introduction (see Fig.~\ref{empirical_result}), assigning a fixed number of communication connections to all clients in a heterogeneous environment is a suboptimal strategy. The results intuitively demonstrate that granting more connections to clients with higher information complexity (i.e., those with a more diverse label distribution) can effectively enhance federated learning performance. This observation constitutes the first key insight for our adaptive topology design: a client's communication needs, reflected by its connectivity degree, should be determined by the information complexity of its local data. This motivation led to the design of the Weighted Label Spatial Dispersion (WLSD) as a universal, automatically computable metric to precisely quantify the information complexity inherent in a client's graph data.

 \noindent \textbf{Weighted Label Spatial Dispersion}. As previously mentioned, this metric provides a comprehensive assessment of a client's intrinsic heterogeneity by quantifying both the structural dispersion and the class diversity of the label distribution within its local subgraph. For a client's graph $\mathcal{G}=(\mathcal{V},\mathcal{E})$, we consider the set of nodes sharing the same label as a sub-population. For each label class $k \in \{1, 2, \dots, K\}$, we define its corresponding set of nodes as $\mathcal{V}_k \subseteq \mathcal{V}$ and calculate the average shortest path distance, $D_k$, between pairs of nodes within this class, computed as:

\begin{equation}
\label{eq:Dk}
D_k = \frac{1}{|\mathcal{V}_k| (|\mathcal{V}_k| - 1)} \sum_{\substack{i, j \in \mathcal{V}_k \\ i \ne j}} d(i, j)
\end{equation}

where $d(i,j)$ denotes the length of the shortest path from node $i$ to node $j$ in the graph, and $|\mathcal{V}_k|$ is the number of nodes in class $k$. This metric measures the degree of dispersion of same-labeled nodes within the graph's structure, reflecting the structural communicability of that label class.

To mitigate the computational bias caused by an excessive or insufficient number of nodes in cases of extreme distribution, we introduce class weights as follows:
\begin{equation}
\label{eq:omega_k}
\omega_k = \frac{\log(1 + |\mathcal{V}_k|)}{\sum_{j = 1}^K \log(1 + |\mathcal{V}_j|)}
\end{equation}

Finally, the weighted average of the structural dispersion is defined as:

\begin{equation}
\label{eq:wlsd_structure}
\mathrm{WLSD} = \sum^{K}_{k = 1} \omega_k \cdot D_k
\end{equation}

By calculating the WLSD, we quantify the heterogeneity of each client's local subgraph into a concise scalar value. This value, which comprehensively reflects the client's intrinsic information complexity, serves as a direct basis for adaptively determining its communication requirements. Specifically, a higher WLSD value implies that the client's labels are more widely dispersed across the graph structure and its information is sparser.

\subsection{Class-aware Semantic Embedding}

\noindent\textbf{Motivation}. After determining the ``quantity" of connections for each client via WLSD, the subsequent challenge is to select for communication ``quality"—that is, which clients to connect with. An intuitive approach is to choose partners that are similar to oneself, but the definition of ``similarity" is of paramount importance. Simply comparing label distributions between clients is far from sufficient, as it completely ignores the most central element of graph data: its topological structure \citep{li2023fedgta}. We posit that the essential characteristic of a client lies not merely in the classes of nodes it possesses, but more importantly, in how these different classes of nodes are intertwined and interrelated across the graph's topology. To achieve this deeper and more precise peer matching, we propose Class-wise Semantic Embedding (CSE): a high-dimensional representation capable of transcending simple label statistics to comprehensively characterize a client's intrinsic semantic-structural properties.

 \noindent\textbf{Class-wise Semantic Embedding}. Our objective is to generate a unique `semantic-structural' fingerprint matrix for each client, designed to capture the interrelationships and diffusion patterns among different label classes within the graph topology. Specifically, for each client, we construct a $K \times K$ semantic representation matrix, where each row represents a class-specific semantic vector $\mathbf{C}_k$. For a sampled set of node pairs of class $k$, denoted as $\mathcal{P}_k=\{(i,j)|y_i=y_j=k,i \neq j\}$, this vector is defined as:

\begin{equation}
\label{eq:cse_k}
\mathbf{C}_k =\frac{1}{|\mathcal{P}_k|} \sum_{(i, j)\in \mathcal{P}_k} \frac{1}{2}(\hat{Y}_{i}+\hat{Y}_{j}) \cdot d(i, j)
\end{equation}

where $\hat{Y}_{i} \in \mathbb{R}^K$ represents the soft label vector of node $i$, $d(i,j)$ is the shortest path length between nodes $i$ and $j$, and $\mathbf{C}_k \in \mathbb{R}^K$ is the structural semantic embedding for class $k$. The final Class-wise Semantic Embedding (CSE) matrix is then formed by concatenating these vectors:

\begin{equation}
\label{eq:cse}
\mathbf{CSE} = \mathbf{C}_1 \parallel\dots\parallel\mathbf{C}_K
\end{equation}

where $\cdot\parallel\cdot$ denotes concatenation. The resulting matrix $\mathbf{CSE} \in \mathbb{R}^{K \times K}$ is the client's final CSE matrix, providing a more comprehensive and profound depiction of its intrinsic heterogeneity and laying a solid foundation for computing inter-client similarity in the subsequent step.

\begin{algorithm}[t]
\caption{The DFed-SST Algorithm}
\label{alg:dfed-sst-algorithm}
\begin{algorithmic}[1]
\Require Clients $\{C_1, \dots, C_N\}$, data $\{G_i\}$, rounds $T$, local epochs $E$, topology update frequency $K_{topo}$.
\Ensure Final models $\{w_1^T, \dots, w_N^T\}$.

\State Initialize $w_i^0$ and initial neighbor set $\mathcal{N}_i$ (e.g., random) for all clients $i$.

\For{each communication round $t = 0, \dots, T-1$}
    \ForAll{client $i \in \{1, \dots, N\}$ }
        \State $\hat{w}_i^{t+1} \gets \text{Perform } E \text{ local update epochs on } w_i^t$.


        \State Receive models $\{\hat{w}_j^{t+1} \mid j \in \mathcal{N}_i\}$ from in-neighbors.
        \State Compute weights $\{\alpha_{ij} \mid j \in \mathcal{N}_i\}$ via Eq. ~\ref{eq:agg_weight}.
        \State $w_i^{t+1} \leftarrow \sum_{j \in \mathcal{N}_i} \alpha_{ij} \hat{w}_j^{t+1}$.\Comment{Perform  aggregation using the current topology $\mathcal{N}_i$.}
    \EndFor
    \If{$t \pmod{K_{topo}} = 0$}\Comment{Periodically update communication topology}
        \State Compute  $\mathrm{WLSD}_i$ \& $\mathrm{CSE}_i$ using $\hat{w}_i^{t+1}$ via Eq.~\ref{eq:wlsd_structure} and Eq. ~\ref{eq:cse}.
  
        \ForAll{client $i \in \{1, \dots, N\}$ }\Comment{Clients select new neighbors.}
            \State Determine in-degree $d_i$ via Eq. ~\ref{eq:degree}.
            \State Compute similarities $S(i,j)$ with all other clients $j$ via Eq.~\ref{eq:similarity}.
            \State Update neighbor set: $\mathcal{N}_i \gets \text{top-}d_i\text{ clients based on } S(i,j)$.
        \EndFor
    \EndIf
\EndFor
\end{algorithmic}
\end{algorithm}

\subsection{Communication Topology Optimization}

 \noindent\textbf{Motivation}. After calculating the heterogeneity metric (WLSD) and the feature fingerprint (CSE) for each client, we possess the foundational components for constructing an adaptive communication network. In DFGL, the rational design of the communication structure is of decisive importance to model convergence speed and generalization ability. Our core motivation is to discard traditional static or random connection methods and propose a communication mechanism entirely and jointly driven by the intrinsic semantic and structural characteristics of the clients. This mechanism not only automatically and differentially determines the ``quantity" of connections (connectivity degree) for each client but also precisely matches them with high-quality communication partners (neighbors), and on this basis, performs a more efficient weighted model aggregation.

 \noindent\textbf{Communication Topology Optimization} We break down the topology optimization process into a three-step procedure that is dynamically executed in each communication round.

The first step is the adaptive determination of communication connectivity. Each client $i$ utilizes its own $\mathrm{WLSD}_i$ value and compares it with the WLSD values of all other clients to determine the number of neighbors it needs to connect with in the current round, i.e., its in-degree, $d_i$. This value is calculated as follows:

\begin{equation}
\label{eq:degree}
d_i = \left| \left\{ j  \ne i\middle|\, \mathrm{WLSD}_j < \mathrm{WLSD}_i \right\} \right|
\end{equation}

This rule embodies an `information potential' principle: a client with a higher WLSD value, representing a more complex internal knowledge structure or sparser information, acts as an ``information seeker." It therefore needs to establish connections with more clients that are relatively ``information-dense" (i.e., those with lower WLSD values) to absorb a broader range of external knowledge.

The second step is the precise selection of communication neighbors. After determining the required number of connections $d_i$, client $i$ must identify the $d_i$ most relevant partners from all other clients. We use the client's CSE matrix to accomplish this task. Specifically, for any other client $j$, we calculate its structural semantic similarity to client $i$ as:

\begin{equation}
\label{eq:similarity}
S(i, j) = \frac{ \langle \mathrm{vec}(\operatorname{CSE}_i), \mathrm{vec}(\operatorname{CSE}_j) \rangle}{\|\mathrm{vec}(\operatorname{CSE}_i)\|_2 \cdot \|\mathrm{vec}(\operatorname{CSE}_j)\|_2}
\end{equation}

Subsequently, client $i$ ranks all other clients in descending order based on the $S(i,j)$ value and selects the top-$d_i$ clients to form its set of in-neighbors, $\mathcal{N}_i$, for the current round. The communication topology constructed through this process is dynamic, asymmetric, and deeply heterogeneity-aware.

The third step is guided weighted model aggregation. After constructing the topology and receiving model updates from its neighbor set $\mathcal{N}_i$, client $i$ performs a weighted model aggregation. To maximize the efficiency of knowledge absorption, we designed an adaptive aggregation weight that fuses similarity with neighbor complexity as follows:

\begin{equation}
\label{eq:agg_weight}
\alpha_{ij} = \frac{\exp(S(i, j)) \cdot \mathrm{WLSD}_j}{\sum_{k \in \mathcal{N}_i} \exp(S(i, k)) \cdot \mathrm{WLSD}_k}
\end{equation}

The intuition behind this weighting strategy is that client $i$, when aggregating the model from its neighbor $j$, will assign a higher weight if neighbor $j$ simultaneously meets two conditions: (1) it has high semantic-structural similarity with client $i$ (i.e., a high $S(i,j)$ value); and (2) it is itself a client with a high degree of intrinsic heterogeneity (i.e., a high $\mathrm{WLSD}_j$ value). Through this strategy, each client can selectively absorb model information from neighbors that are structurally more comprehensive and semantically more relevant during the local aggregation process, thereby enhancing personalization and boosting overall system performance. The complete algorithm is presented in Algorithm~\ref{alg:dfed-sst-algorithm}.

\begin{table}[t]
    \centering
    \caption{The statistical information of the experimental datasets.}
    \label{datasetSplit}
    \resizebox{\textwidth}{!}{
    \begin{tabular}{c|ccccccc}
    \hline
        \textbf{dataset} & \textbf{nodes} & \textbf{features} & \textbf{edges} & \textbf{classes} & \textbf{train/val/test} & \textbf{description} \\ \hline
        Cora & 2,708 & 1,433 & 5,429 & 7 & 20\%/40\%/40\% & citation network \\ 
        CiteSeer & 3,327 & 3,703 & 4,732 & 6 & 20\%/40\%/40\% & citation network \\ 
        PubMed & 19,717 & 500 & 44,338 & 3 & 20\%/40\%/40\% & citation network \\ \hline
        Amazon Photo & 7,487 & 745 & 119,043 & 8 & 20\%/40\%/40\% & co-purchase graph \\ 
        Amazon Computer & 13,381 & 767 & 245,778 & 10 & 20\%/40\%/40\% & co-purchase graph \\ \hline
        Coauthor CS & 18,333 & 6,805 & 81,894 & 15 & 20\%/40\%/40\% & co-authorship graph \\ 
        Coauthor Physics & 34,493 & 8,415 & 247,962 & 5 & 20\%/40\%/40\% & co-authorship graph \\ \hline
        ogbn-arxiv & 169,343 & 128 & 2,315,598 & 40 & 60\%/20\%/20\% & citation network \\ \hline
    \end{tabular}}
\end{table}
\section{Experiments}
\label{Experiments}
In this section, we conduct a series of experiments to validate the effectiveness of our proposed method. Our objective is to answer the following research questions: \textbf{Q1:} How does our method perform compared to state-of-the-art decentralized methods? \textbf{Q2:} What is the contribution of each key component in our method? \textbf{Q3:} How robust is our method to data sparsity? \textbf{Q4:} What is the computational efficiency of our method?

\subsection{Experimental Setup}
\begin{table*}[t]
\renewcommand{\arraystretch}{1.2} 
\centering
\caption{Performance comparison of test accuracy (\%) achieved by DFed-SST and baseline methods on eight datasets. The best results are highlighted in \textbf{bold}.}
\label{tab:main_results_comparison}
\resizebox{\textwidth}{!}{%
\begin{tabular}{l|cc cc cc cc}
\toprule
Dataset $(\rightarrow)$ & \multicolumn{2}{c}{Cora} & \multicolumn{2}{c}{CiteSeer} & \multicolumn{2}{c}{PubMed} & \multicolumn{2}{c}{Computers} \\
\cline{2-3} \cline{4-5} \cline{6-7} \cline{8-9}
\rule{0pt}{1.1em}Method $(\downarrow)$ & 10 Clients & 20 Clients & 10 Clients & 20 Clients & 10 Clients & 20 Clients & 10 Clients & 20 Clients \\
\midrule
Gossip   & $79.97 \pm 0.5$ & $74.94 \pm 0.7$ & $66.94 \pm 0.4$ & $65.81 \pm 0.4$ & $84.88 \pm 0.1$ & $84.64 \pm 0.2$ & $83.53 \pm 0.6$ & $81.41 \pm 1.0$ \\
PENS     & $79.95 \pm 0.3$ & $75.78 \pm 0.4$ & $69.97 \pm 0.0$ & $67.19 \pm 0.2$ & $85.10 \pm 0.2$ & $84.67 \pm 0.1$ & $84.16 \pm 0.7$ & $82.84 \pm 0.2$ \\
DFedAvgM & $77.41 \pm 0.7$ & $68.74 \pm 0.3$ & $58.97 \pm 0.6$ & $59.93 \pm 0.5$ & $76.95 \pm 0.5$ & $77.74 \pm 0.2$ & $80.52 \pm 0.7$ & $80.78 \pm 0.8$ \\
FedDual  & $78.98 \pm 0.2$ & $74.21 \pm 0.7$ & $67.38 \pm 0.6$ & $63.57 \pm 0.9$ & $84.29 \pm 0.3$ & $83.97 \pm 0.3$ & $83.29 \pm 1.8$ & $81.73 \pm 0.5$ \\
D-PSGD   & $78.32 \pm 0.3$ & $72.89 \pm 0.5$ & $68.78 \pm 0.2$ & $65.55 \pm 0.3$ & $82.20 \pm 0.2$ & $81.83 \pm 0.2$ & $81.65 \pm 0.7$ & $79.10 \pm 0.5$ \\
OSGP     & $79.10 \pm 0.1$ & $74.21 \pm 0.4$ & $68.87 \pm 0.4$ & $67.63 \pm 0.1$ & $84.89 \pm 0.0$ & $84.69 \pm 0.0$ & $84.65 \pm 0.7$ & $79.55 \pm 0.4$ \\
Dis-PFL  & $78.86 \pm 0.4$ & $74.89 \pm 0.3$ & $67.58 \pm 0.2$ & $64.52 \pm 0.3$ & $83.75 \pm 0.3$ & $83.15 \pm 0.4$ & $82.91 \pm 0.8$ & $81.19 \pm 0.6$ \\
\textbf{DFed-SST (Ours)} & \bm{$81.16 \pm 0.2$} & \bm{$76.97 \pm 0.4$} & \bm{$70.48 \pm 0.1$} & \bm{$69.41 \pm 0.2$} & \bm{$86.40 \pm 0.1$} & \bm{$86.27 \pm 0.0$} & \bm{$87.90 \pm 0.5$} & \bm{$84.17 \pm 0.4$} \\
\bottomrule
\toprule
Dataset $(\rightarrow)$ & \multicolumn{2}{c}{Physics} & \multicolumn{2}{c}{CS} & \multicolumn{2}{c}{Photo} & \multicolumn{2}{c}{ogbn-arxiv} \\
\cline{2-3} \cline{4-5} \cline{6-7} \cline{8-9}
\rule{0pt}{1.1em}Method $(\downarrow)$ & 10 Clients & 20 Clients & 10 Clients & 20 Clients & 10 Clients & 20 Clients & 10 Clients & 20 Clients \\
\midrule
Gossip   & $94.34 \pm 0.1$ & $93.55 \pm 0.1$ & $89.33 \pm 0.2$ & $86.98 \pm 0.1$ & $86.88 \pm 1.1$ & $85.43 \pm 0.6$ & $66.12 \pm 0.3$ & $63.85 \pm 0.2$ \\
PENS     & $94.37 \pm 0.0$ & $93.71 \pm 0.1$ & $89.39 \pm 0.1$ & $87.23 \pm 0.2$ & $90.62 \pm 0.4$ & $87.04 \pm 0.7$ & $66.33 \pm 0.1$ & $64.74 \pm 0.2$ \\
DFedAvgM & $87.99 \pm 0.2$ & $89.61 \pm 0.0$ & $80.92 \pm 0.2$ & $80.42 \pm 0.1$ & $85.65 \pm 0.4$ & $85.50 \pm 0.5$ & $62.61 \pm 0.5$ & $58.36 \pm 0.2$ \\
FedDual  & $94.19 \pm 0.0$ & $93.24 \pm 0.1$ & $87.24 \pm 0.1$ & $86.06 \pm 0.2$ & $87.46 \pm 0.8$ & $86.06 \pm 0.2$ & $63.83 \pm 0.3$ & $62.87 \pm 0.4$ \\
D-PSGD   & $94.09 \pm 0.1$ & $93.66 \pm 0.1$ & $89.06 \pm 0.1$ & $86.61 \pm 0.1$ & $85.08 \pm 0.7$ & $85.18 \pm 0.3$ & $65.73 \pm 0.6$ & $62.19 \pm 0.8$ \\
OSGP     & $94.27 \pm 0.1$ & $93.62 \pm 0.1$ & $88.13 \pm 0.1$ & $87.31 \pm 0.1$ & $89.29 \pm 0.2$ & $86.63 \pm 0.6$ & $60.32 \pm 0.1$ & $56.85 \pm 0.2$ \\
Dis-PFL  & $94.15 \pm 0.1$ & $93.48 \pm 0.1$ & $88.54 \pm 0.2$ & $86.37 \pm 0.2$ & $87.88 \pm 0.5$ & $86.21 \pm 0.4$ & $64.21 \pm 0.4$ & $63.05 \pm 0.3$ \\
\textbf{DFed-SST (Ours)} & \bm{$94.40 \pm 0.1$} & \bm{$94.25 \pm 0.1$} & \bm{$90.47 \pm 0.1$} & \bm{$89.11 \pm 0.1$} & \bm{$91.89 \pm 0.2$} & \bm{$91.22 \pm 0.1$} & \bm{$66.96 \pm 0.1$} & \bm{$64.94 \pm 0.1$} \\
\bottomrule
\end{tabular}%
}
\end{table*}
\noindent{\textbf{Datasets}.}
We conduct experiments on eight widely-used public benchmark datasets for graph learning: three small-scale citation networks (Cora, CiteSeer, PubMed) \citep{yang2016revisiting}; two medium-scale user-item datasets (Amazon Computer, Amazon Photo); two medium-scale co-author datasets (Coauthor CS, Coauthor Physics) \citep{shchur2018pitfalls}; and one large-scale OGB dataset (ogbn-arxiv)~\cite{hu2020open}. On these datasets, we use the Metis algorithm to partition and assign nodes to a given number of clients. This partitioning scheme is widely used in FGL to simulate distributed scenarios in subgraph-based federated learning. Detailed descriptions of these datasets are presented in Table~\ref{datasetSplit}.

\begin{figure}[t]
 \centering
 \includegraphics[width=0.9\textwidth]{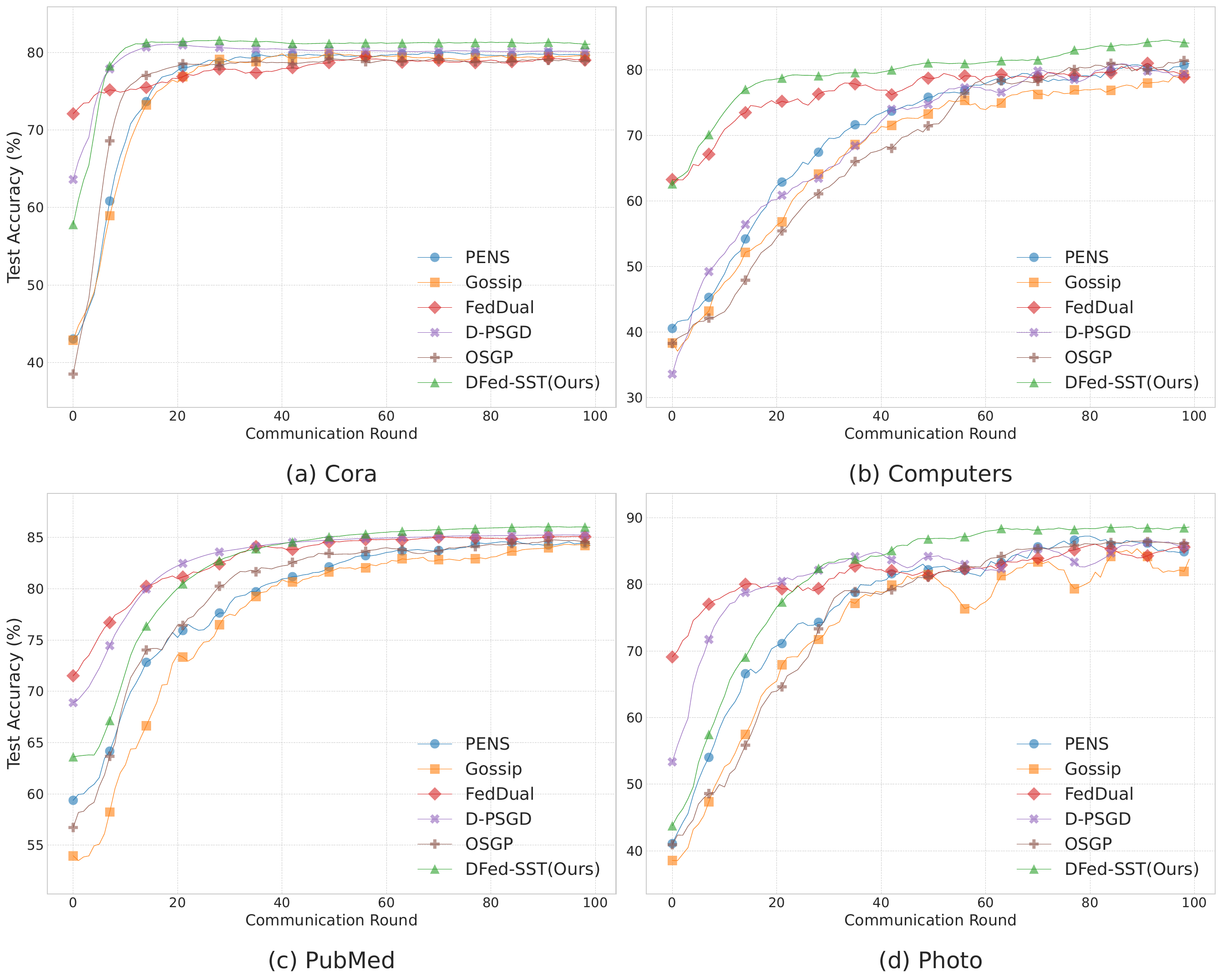}%
 \caption{Convergence curves of our proposed DFed-SST and baseline methods on four graph datasets with 10 participating clients.}
 \label{curves}
 \vspace{-10pt}
\end{figure}

\noindent\textbf{Baselines.} We compare our method against seven decentralized federated learning methods, including four traditional decentralized methods (DFedAvgM \citep{sun2022decentralized}, OSGP \citep{assran2019stochastic}, Dis-PFL \citep{dai2022dispfl}, D-PSGD \citep{lian2017can}), two gossip-based learning methods (Gossip \citep{hu2019decentralized}, FedDual \citep{chen2022feddual}), and one method with an adaptive topology (PENS \citep{onoszko2021decentralized}). A Graph Convolutional Network (GCN) serves as the backbone architecture for all tested methods.

\noindent\textbf{Hyperparameters and Experimental Settings.} For the baseline methods, we adopt the hyperparameter settings reported in their original papers. Unless otherwise specified, we use a consistent configuration for all algorithms. We employ a two-layer GCN as the base model, with a hidden dimension of 64 and a learning rate of 1e-2. We set the number of local training epochs to 3 and the number of communication rounds to 100. For the communication topology, we adopt a dynamic, time-varying connectivity model. For each experiment, we report the mean and variance over five independent runs.

\noindent\textbf{Experiment Environment.} We conduct our experiments on a computer equipped with an Intel Core i7-13700K CPU @ 3.40 GHz, an NVIDIA GeForce RTX 3090, and 64 GB of memory running CUDA 12.6. The operating system is Ubuntu 22.04.

\subsection{Performance Comparison}

To answer \textbf{Q1}, we conducted experiments on multiple datasets, partitioned into scenarios with 10 and 20 clients using the Metis scheme. For each baseline, we performed five training runs and report the mean accuracy and its variance. Our proposed method, DFed-SST, consistently outperforms the baselines. Specifically, compared to Gossip, DFed-SST yields performance gains of up to 5.79\%; compared to PENS, the strongest baseline in most cases, DFed-SST achieves a performance improvement of up to 4.18\%. The detailed results are presented in Table~\ref{tab:main_results_comparison}. The convergence curves for DFed-SST and the baselines are displayed in Fig.~\ref{curves}.
\begin{figure}[t]
\rmfamily
\centering
\includegraphics[width=0.9\textwidth]{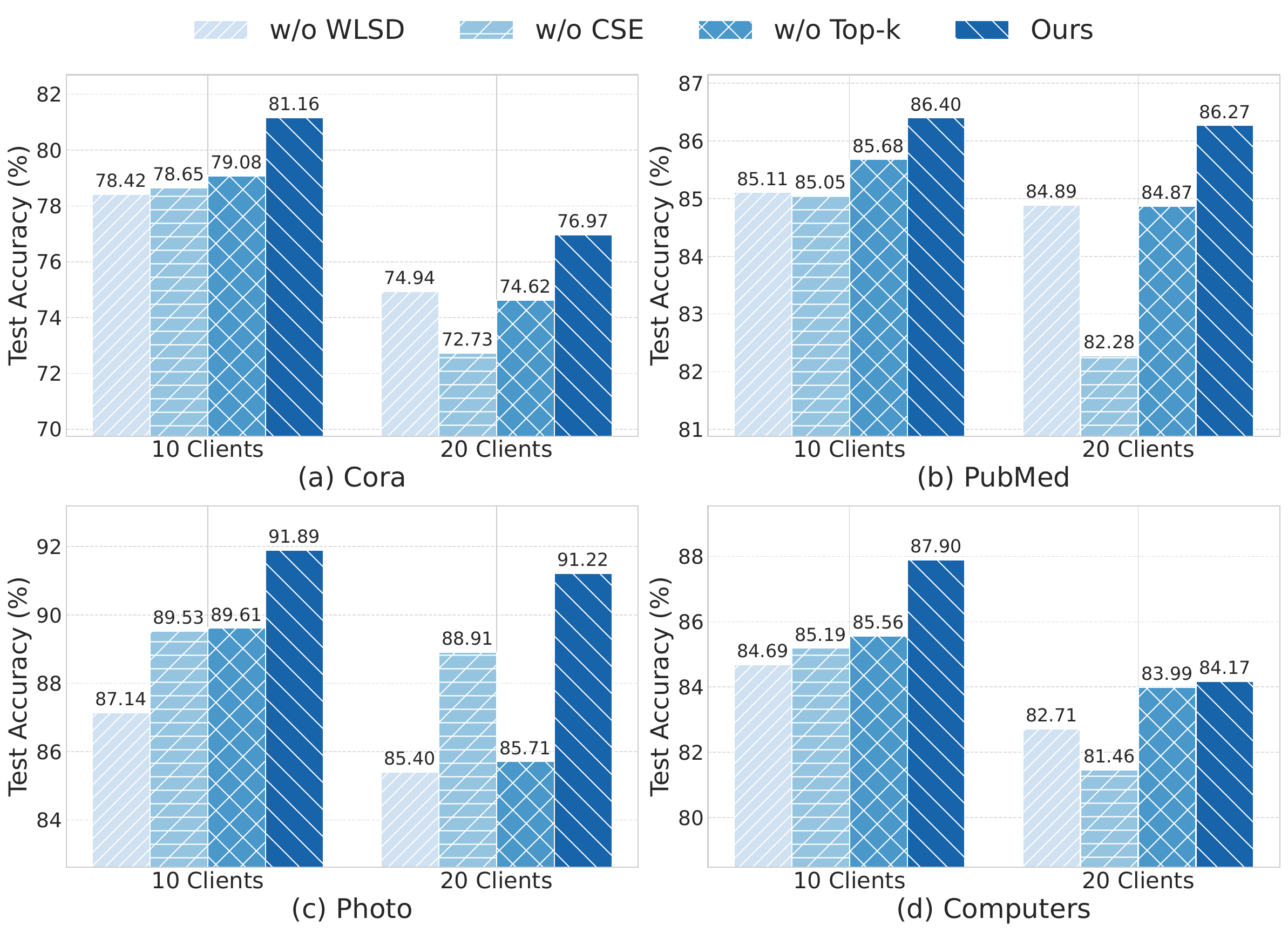} 

\DeclareGraphicsExtensions.
\caption{Experimental results for the ablation study.}
\label{abl}
\end{figure}

\begin{figure}[t]
\rmfamily
\centering

\subfigure[Cora]{
\includegraphics[width=0.92\textwidth]{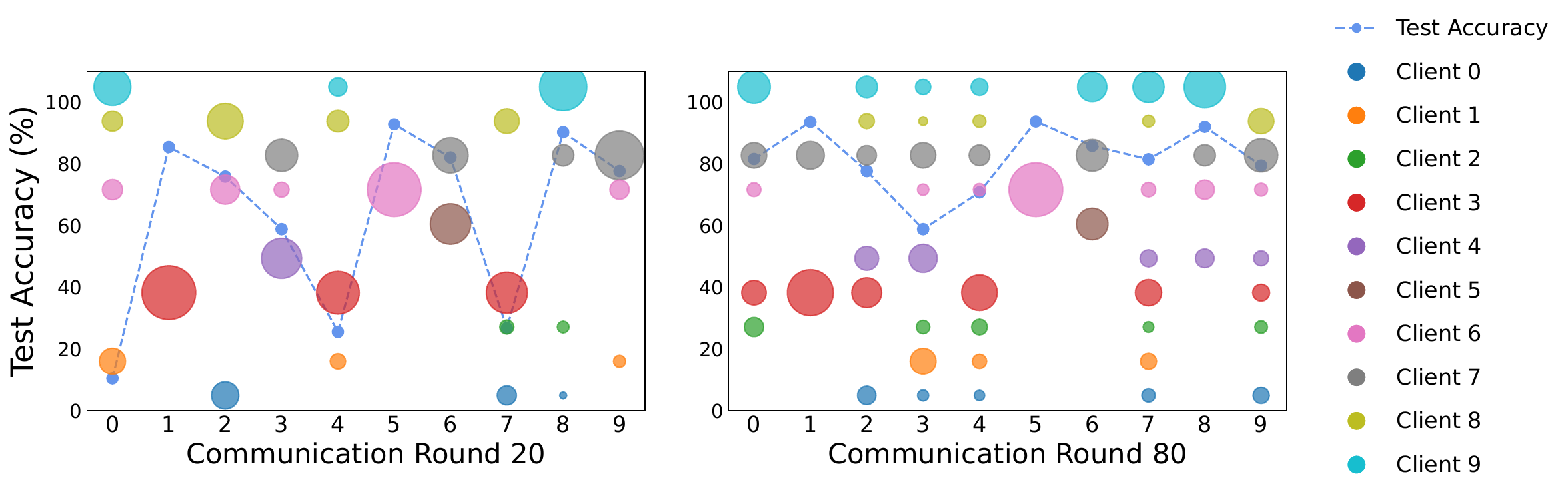}
\label{fig:cora_visual}
}
\subfigure[Photo]{
\includegraphics[width=0.92\textwidth]{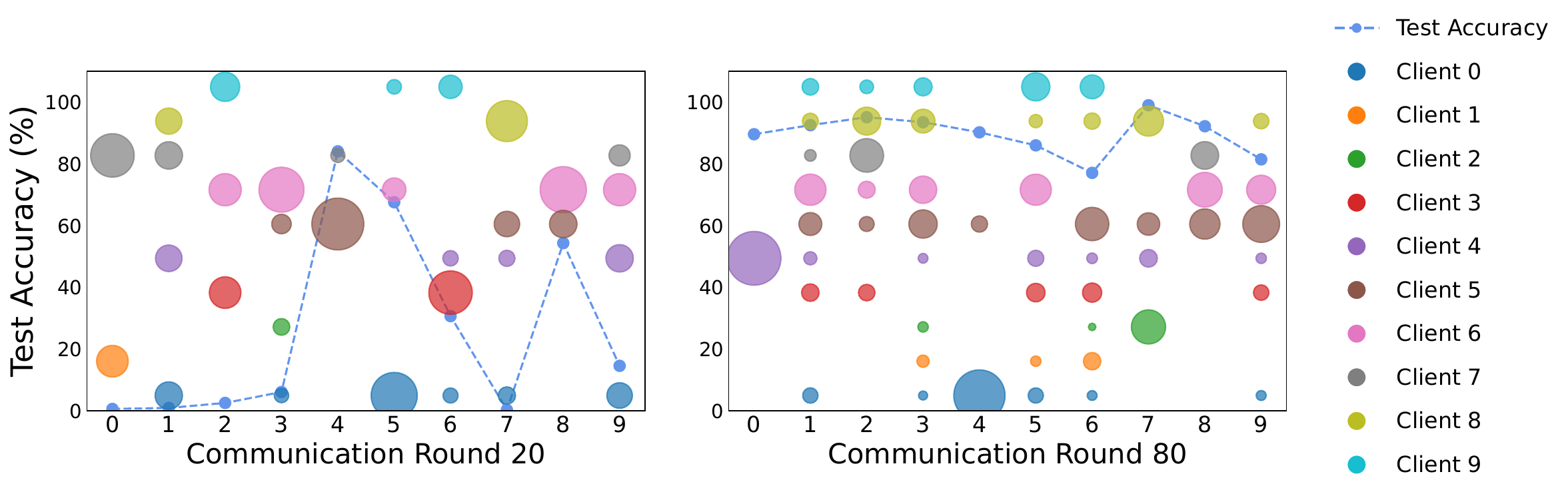} 
\label{fig:photo_visual}
}

\caption{Visualization of the communication topology on the Photo and Cora datasets across 10 clients. Circles of different colors represent different clients, and their sizes indicate the model aggregation weights.}
\label{topo_visual}
\end{figure}

\begin{table*}[t]
\centering
\caption{Performance comparison of test accuracy (\%) under Random Topology Sparsity.}
\label{tab:topology_sparsity}
\resizebox{\textwidth}{!}{%
\begin{tabular}{lcccccccc}
\toprule
\textbf{Method} & \textbf{Cora} & \textbf{CiteSeer} & \textbf{PubMed} & \textbf{Computers} & \textbf{Physics} & \textbf{CS} & \textbf{Photo} & \textbf{ogbn-arxiv} \\
\midrule
Gossip          & $78.01 \pm 0.7$ & $66.94 \pm 0.4$ & $84.88 \pm 0.1$ & $82.25 \pm 0.9$ & $94.34 \pm 0.1$ & $89.33 \pm 0.2$ & $86.88 \pm 1.1$ & $66.12 \pm 0.3$ \\
PENS            & $78.67 \pm 0.1$ & $69.97 \pm 0.0$ & $85.10 \pm 0.2$ & $83.40 \pm 0.8$ & $94.37 \pm 0.0$ & $89.39 \pm 0.1$ & $90.62 \pm 0.4$ & $66.33 \pm 0.1$ \\
DFedAvgM        & $75.51 \pm 0.8$ & $55.23 \pm 0.9$ & $80.11 \pm 0.5$ & $78.45 \pm 1.0$ & $90.53 \pm 0.4$ & $83.15 \pm 0.6$ & $82.05 \pm 0.9$ & $61.82 \pm 0.7$ \\
FedDual         & $77.15 \pm 0.6$ & $63.88 \pm 0.5$ & $83.24 \pm 0.4$ & $81.06 \pm 0.8$ & $92.88 \pm 0.3$ & $87.21 \pm 0.4$ & $85.93 \pm 0.6$ & $64.11 \pm 0.5$ \\
D-PSGD          & $77.89 \pm 0.4$ & $65.17 \pm 0.3$ & $84.05 \pm 0.3$ & $81.73 \pm 0.7$ & $93.95 \pm 0.2$ & $88.66 \pm 0.3$ & $86.11 \pm 0.5$ & $65.34 \pm 0.4$ \\
OSGP            & $78.12 \pm 0.3$ & $66.29 \pm 0.2$ & $84.51 \pm 0.2$ & $82.01 \pm 0.6$ & $94.18 \pm 0.1$ & $89.02 \pm 0.2$ & $86.57 \pm 0.4$ & $65.88 \pm 0.2$ \\
Dis-PFL         & $77.86 \pm 0.5$ & $58.97 \pm 0.6$ & $84.55 \pm 0.4$ & $81.90 \pm 0.7$ & $94.15 \pm 0.2$ & $88.75 \pm 0.3$ & $88.10 \pm 0.8$ & $65.50 \pm 0.4$ \\
\textbf{DFed-SST (Ours)} & $\mathbf{80.43 \pm 0.3}$ & $\mathbf{70.72 \pm 0.2}$ & $\mathbf{85.74 \pm 0.09}$ & $\mathbf{87.30 \pm 0.6}$ & $\mathbf{94.40 \pm 0.1}$ & $\mathbf{90.47 \pm 0.1}$ & $\mathbf{91.14 \pm 0.5}$ & $\mathbf{66.96 \pm 0.1}$ \\
\bottomrule
\end{tabular}%
}
\end{table*}

\begin{table*}[t]
\centering
\caption{Performance comparison of test accuracy (\%) under Random Label Sparsity.}
\label{tab:label_sparsity}
\resizebox{\textwidth}{!}{%
\begin{tabular}{lcccccccc}
\toprule
\textbf{Method} & \textbf{Cora} & \textbf{CiteSeer} & \textbf{PubMed} & \textbf{Computers} & \textbf{Physics} & \textbf{CS} & \textbf{Photo} & \textbf{ogbn-arxiv} \\
\midrule
Gossip          & $78.15 \pm 0.6$ & $67.05 \pm 0.4$ & $85.05 \pm 0.2$ & $82.35 \pm 0.8$ & $94.28 \pm 0.2$ & $89.25 \pm 0.3$ & $87.05 \pm 1.0$ & $66.25 \pm 0.4$ \\
PENS            & $78.72 \pm 0.2$ & $69.85 \pm 0.1$ & $85.25 \pm 0.1$ & $83.50 \pm 0.7$ & $94.35 \pm 0.1$ & $89.45 \pm 0.2$ & $90.55 \pm 0.5$ & $66.40 \pm 0.2$ \\
DFedAvgM        & $75.30 \pm 0.9$ & $56.10 \pm 0.8$ & $80.50 \pm 0.6$ & $78.80 \pm 1.1$ & $90.30 \pm 0.5$ & $83.50 \pm 0.7$ & $82.50 \pm 0.8$ & $62.10 \pm 0.6$ \\
FedDual         & $78.95 \pm 0.3$ & $68.90 \pm 0.3$ & $85.45 \pm 0.2$ & $83.15 \pm 0.6$ & $94.10 \pm 0.2$ & $88.90 \pm 0.3$ & $88.55 \pm 0.4$ & $65.95 \pm 0.3$ \\
D-PSGD          & $77.50 \pm 0.5$ & $65.50 \pm 0.4$ & $84.10 \pm 0.4$ & $81.50 \pm 0.8$ & $93.50 \pm 0.3$ & $88.10 \pm 0.4$ & $86.20 \pm 0.6$ & $65.10 \pm 0.5$ \\
OSGP            & $77.80 \pm 0.4$ & $66.80 \pm 0.3$ & $84.60 \pm 0.3$ & $82.10 \pm 0.7$ & $93.80 \pm 0.2$ & $88.50 \pm 0.3$ & $86.80 \pm 0.5$ & $65.50 \pm 0.4$ \\
Dis-PFL         & $78.50 \pm 0.4$ & $70.15 \pm 0.2$ & $85.15 \pm 0.3$ & $82.80 \pm 0.5$ & $94.05 \pm 0.1$ & $89.15 \pm 0.2$ & $89.05 \pm 0.3$ & $66.75 \pm 0.2$ \\
\textbf{DFed-SST (Ours)} & $\mathbf{80.65 \pm 0.2}$ & $\mathbf{70.80 \pm 0.1}$ & $\mathbf{85.95 \pm 0.1}$ & $\mathbf{87.40 \pm 0.4}$ & $\mathbf{94.45 \pm 0.1}$ & $\mathbf{90.55 \pm 0.1}$ & $\mathbf{91.30 \pm 0.2}$ & $\mathbf{67.15 \pm 0.1}$ \\
\bottomrule
\end{tabular}%
}
\end{table*}

\subsection{Method Interpretability}

To answer \textbf{Q2}, we conduct an ablation study to investigate the contribution of each key module in our proposed DFed-SST framework. We design the following three variants, each ablating a specific component: \textbf{1. Without WLSD-based Degree Control}, where we replace our WLSD-based mechanism with a fixed connection degree (half the total number of clients) for each client; \textbf{2. Without CSE for Similarity}, where the CSE-based similarity metric is replaced by one calculated from the mean of node features; and \textbf{3. Without Similarity-based Selection}, where the Top-K neighbor selection mechanism is replaced with random neighbor selection. As observed in Fig.~\ref{abl}, removing any of these components from our communication topology optimization process results in performance degradation.

To further validate the effectiveness of our adaptive topology optimization, we visualized the communication topologies at different communication rounds during a single training process, with the results presented in Fig.~\ref{topo_visual}. The visualizations reveal a dynamic evolutionary trend as training progresses, where clients continuously adjust their communication partners in accordance with their changing data characteristics. This dynamic optimization process is strongly correlated with the steady improvement in overall model accuracy, intuitively demonstrating our method's ability to construct an efficient and continuously evolving communication network.

\subsection{Robustness Analysis}
To answer \textbf{Q3}, we conduct two sets of experiments on all graph datasets to evaluate our method's robustness to data sparsity. We introduce two types of perturbations: \textbf{random label sparsity} and \textbf{random topological structure sparsity}. Each baseline model was trained for five runs, and we report the mean performance and its variance. As presented in Table~\ref{tab:label_sparsity} and Table~\ref{tab:topology_sparsity}, the results demonstrate that our proposed DFed-SST consistently achieves the best test accuracy across all datasets, under both sparse local topology and sparse label conditions. This outcome validates the robustness of DFed-SST against data sparsity.

\subsection{Efficiency  Analysis}

\begin{figure}
 \centering
 \includegraphics[width=0.95\textwidth]{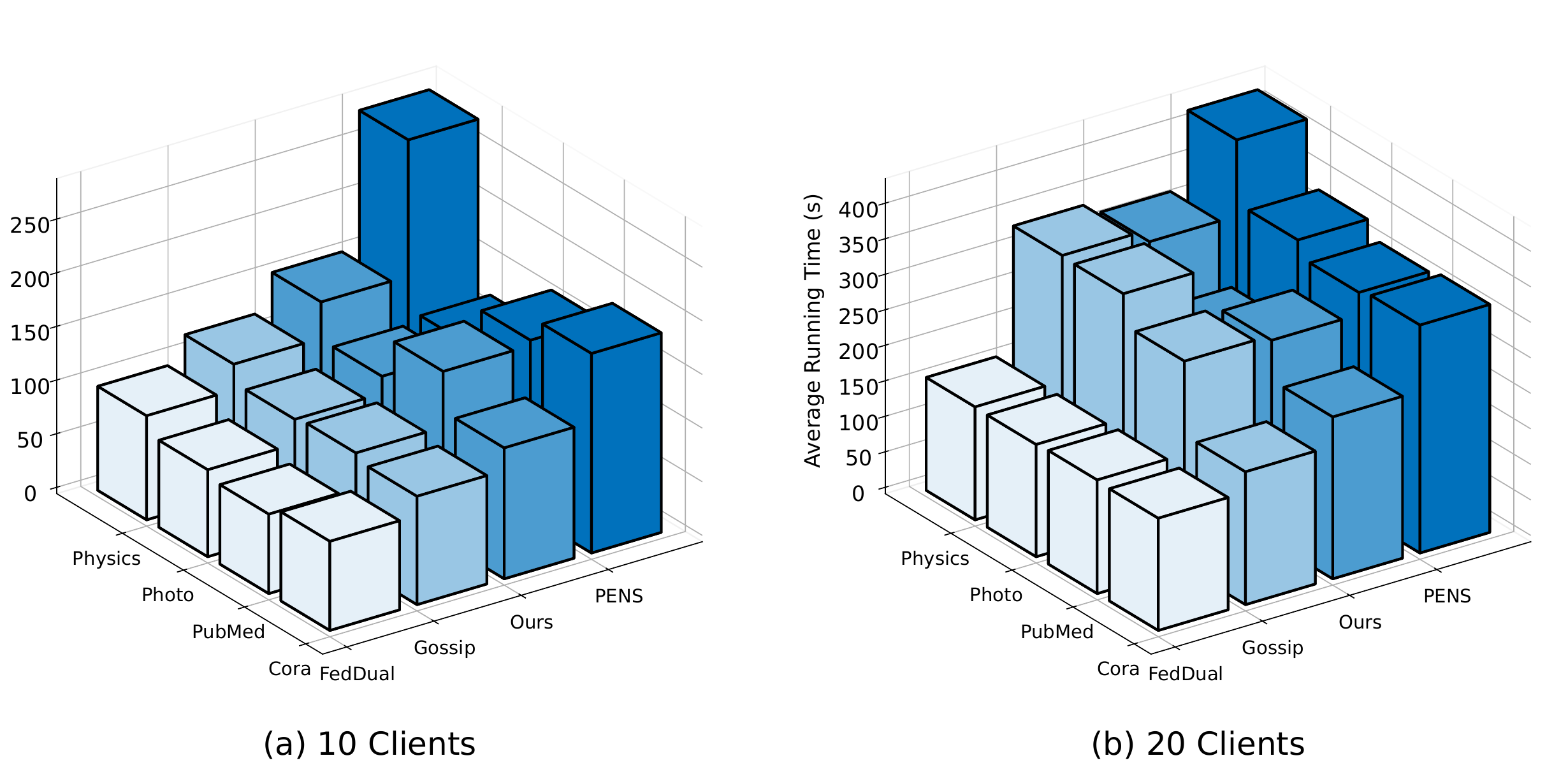}%
 \caption{Comparison of average per-round running time across four datasets under 10- and 20-client scenarios.}
 \label{running_time}
 \vspace{-10pt}
\end{figure}

To answer \textbf{Q4}, we designed an efficiency comparison experiment to measure and compare the average running time of DFed-SST against three other baseline methods over multiple rounds. We conducted these experiments on four datasets under settings with 10 and 20 clients, respectively. The results are shown in Fig.~\ref{running_time}. FedDual exhibits the highest efficiency due to its pairwise communication and aggregation design. In contrast, PENS incurs higher computational costs because it requires evaluating received models locally. Although our method involves additional computations, we adjust the communication topology periodically rather than in every round. Therefore, compared to Gossip, our method shows no significant overhead. Furthermore, as the number of clients increases, the growth in its computational cost is also lower than that of Gossip, striking an effective balance between computational overhead and performance gains.

\section{Conclusion}
\label{Conclusion}
In this paper, we tackle a core challenge in Decentralized Federated Graph Learning (DFGL): the inability of existing communication topologies to adapt to the profound dual semantic and structural heterogeneity presented by clients. To move beyond the limitations of static, random, or structure-agnostic topologies, we propose DFed-SST, a new paradigm for data-driven dynamic topology construction. The framework's core innovation lies in two novel measurement tools, WLSD and CSE, which are the first to simultaneously quantify a client's dual semantic and structural heterogeneity. Leveraging these metrics, our framework adaptively determines the required number of connections and precisely matches each client with its most relevant communication partners. Extensive experiments on eight highly heterogeneous, real-world datasets validate the superiority of our proposed framework. The results show that DFed-SST is significantly superior to various state-of-the-art baseline methods in both model accuracy and convergence efficiency, offering a practical solution for efficient federated graph learning in complex, decentralized environments.

Despite these encouraging results, several avenues for future work remain. First, future research could explore lightweight approximation methods for WLSD and CSE to reduce the computational overhead of topology construction in large-scale networks. Second, a promising direction involves extending the framework to support asynchronous decentralized federated systems, which more closely align with real-world scenarios. Finally, integrating our adaptive topology concept with other federated learning techniques holds the potential to further enhance model performance.

\section*{Data availability}
The data used in this research are publicly available graph datasets.

\clearpage

\bibliographystyle{elsarticle-harv}
\bibliography{main}

\begin{thebibliography}{32}
\expandafter\ifx\csname natexlab\endcsname\relax\def\natexlab#1{#1}\fi
\providecommand{\url}[1]{\texttt{#1}}
\providecommand{\href}[2]{#2}
\providecommand{\path}[1]{#1}
\providecommand{\DOIprefix}{doi:}
\providecommand{\ArXivprefix}{arXiv:}
\providecommand{\URLprefix}{URL: }
\providecommand{\Pubmedprefix}{pmid:}
\providecommand{\doi}[1]{\href{http://dx.doi.org/#1}{\path{#1}}}
\providecommand{\Pubmed}[1]{\href{pmid:#1}{\path{#1}}}
\providecommand{\bibinfo}[2]{#2}
\ifx\xfnm\relax \def\xfnm[#1]{\unskip,\space#1}\fi
\bibitem[{Assran et~al.(2019)Assran, Loizou, Ballas and Rabbat}]{assran2019stochastic}
\bibinfo{author}{Assran, M.}, \bibinfo{author}{Loizou, N.}, \bibinfo{author}{Ballas, N.}, \bibinfo{author}{Rabbat, M.}, \bibinfo{year}{2019}.
\newblock \bibinfo{title}{Stochastic gradient push for distributed deep learning}, in: \bibinfo{booktitle}{International Conference on Machine Learning}, \bibinfo{organization}{PMLR}. pp. \bibinfo{pages}{344--353}.
\bibitem[{Baek et~al.(2023)Baek, Jeong, Jin, Yoon and Hwang}]{baek2023personalized}
\bibinfo{author}{Baek, J.}, \bibinfo{author}{Jeong, W.}, \bibinfo{author}{Jin, J.}, \bibinfo{author}{Yoon, J.}, \bibinfo{author}{Hwang, S.J.}, \bibinfo{year}{2023}.
\newblock \bibinfo{title}{Personalized subgraph federated learning}, in: \bibinfo{booktitle}{International conference on machine learning}, \bibinfo{organization}{PMLR}. pp. \bibinfo{pages}{1396--1415}.
\bibitem[{Beltr{\'a}n et~al.(2023)Beltr{\'a}n, P{\'e}rez, S{\'a}nchez, Bernal, Bovet, P{\'e}rez, P{\'e}rez and Celdr{\'a}n}]{beltran2023decentralized}
\bibinfo{author}{Beltr{\'a}n, E.T.M.}, \bibinfo{author}{P{\'e}rez, M.Q.}, \bibinfo{author}{S{\'a}nchez, P.M.S.}, \bibinfo{author}{Bernal, S.L.}, \bibinfo{author}{Bovet, G.}, \bibinfo{author}{P{\'e}rez, M.G.}, \bibinfo{author}{P{\'e}rez, G.M.}, \bibinfo{author}{Celdr{\'a}n, A.H.}, \bibinfo{year}{2023}.
\newblock \bibinfo{title}{Decentralized federated learning: Fundamentals, state of the art, frameworks, trends, and challenges}.
\newblock \bibinfo{journal}{IEEE Communications Surveys \& Tutorials} \bibinfo{volume}{25}, \bibinfo{pages}{2983--3013}.
\bibitem[{Chen et~al.(2022)Chen, Wang, Wang and Lin}]{chen2022feddual}
\bibinfo{author}{Chen, Q.}, \bibinfo{author}{Wang, Z.}, \bibinfo{author}{Wang, H.}, \bibinfo{author}{Lin, X.}, \bibinfo{year}{2022}.
\newblock \bibinfo{title}{Feddual: Pair-wise gossip helps federated learning in large decentralized networks}.
\newblock \bibinfo{journal}{IEEE Transactions on Information Forensics and Security} \bibinfo{volume}{18}, \bibinfo{pages}{335--350}.
\bibitem[{Dai et~al.(2022)Dai, Shen, He, Tian and Tao}]{dai2022dispfl}
\bibinfo{author}{Dai, R.}, \bibinfo{author}{Shen, L.}, \bibinfo{author}{He, F.}, \bibinfo{author}{Tian, X.}, \bibinfo{author}{Tao, D.}, \bibinfo{year}{2022}.
\newblock \bibinfo{title}{Dispfl: Towards communication-efficient personalized federated learning via decentralized sparse training}.
\newblock \bibinfo{journal}{arXiv preprint arXiv:2206.00187} .
\bibitem[{Ekstr{\"o}m~Kelvinius et~al.(2023)Ekstr{\"o}m~Kelvinius, Georgiev, Toshev and Gasteiger}]{ekstrom2023accelerating}
\bibinfo{author}{Ekstr{\"o}m~Kelvinius, F.}, \bibinfo{author}{Georgiev, D.}, \bibinfo{author}{Toshev, A.}, \bibinfo{author}{Gasteiger, J.}, \bibinfo{year}{2023}.
\newblock \bibinfo{title}{Accelerating molecular graph neural networks via knowledge distillation}.
\newblock \bibinfo{journal}{Advances in Neural Information Processing Systems} \bibinfo{volume}{36}, \bibinfo{pages}{25761--25792}.
\bibitem[{Geng et~al.(2024)Geng, Liu and Huang}]{geng2024privacy}
\bibinfo{author}{Geng, T.}, \bibinfo{author}{Liu, J.}, \bibinfo{author}{Huang, C.T.}, \bibinfo{year}{2024}.
\newblock \bibinfo{title}{A privacy-preserving federated learning framework for iot environment based on secure multi-party computation}, in: \bibinfo{booktitle}{2024 IEEE Annual Congress on Artificial Intelligence of Things (AIoT)}, \bibinfo{organization}{IEEE}. pp. \bibinfo{pages}{117--122}.
\bibitem[{Hu et~al.(2019)Hu, Jiang and Wang}]{hu2019decentralized}
\bibinfo{author}{Hu, C.}, \bibinfo{author}{Jiang, J.}, \bibinfo{author}{Wang, Z.}, \bibinfo{year}{2019}.
\newblock \bibinfo{title}{Decentralized federated learning: A segmented gossip approach}.
\newblock \bibinfo{journal}{arXiv preprint arXiv:1908.07782} .
\bibitem[{Hu et~al.(2020)Hu, Fey, Zitnik, Dong, Ren, Liu, Catasta and Leskovec}]{hu2020open}
\bibinfo{author}{Hu, W.}, \bibinfo{author}{Fey, M.}, \bibinfo{author}{Zitnik, M.}, \bibinfo{author}{Dong, Y.}, \bibinfo{author}{Ren, H.}, \bibinfo{author}{Liu, B.}, \bibinfo{author}{Catasta, M.}, \bibinfo{author}{Leskovec, J.}, \bibinfo{year}{2020}.
\newblock \bibinfo{title}{Open graph benchmark: Datasets for machine learning on graphs}.
\newblock \bibinfo{journal}{Advances in neural information processing systems} \bibinfo{volume}{33}, \bibinfo{pages}{22118--22133}.
\bibitem[{Lalitha et~al.(2018)Lalitha, Shekhar, Javidi and Koushanfar}]{lalitha2018fully}
\bibinfo{author}{Lalitha, A.}, \bibinfo{author}{Shekhar, S.}, \bibinfo{author}{Javidi, T.}, \bibinfo{author}{Koushanfar, F.}, \bibinfo{year}{2018}.
\newblock \bibinfo{title}{Fully decentralized federated learning}, in: \bibinfo{booktitle}{Third workshop on bayesian deep learning (NeurIPS)}.
\bibitem[{Li et~al.(2025a)Li, Yu, Xia and Pang}]{li2025centralized}
\bibinfo{author}{Li, Q.}, \bibinfo{author}{Yu, W.}, \bibinfo{author}{Xia, Y.}, \bibinfo{author}{Pang, J.}, \bibinfo{year}{2025}a.
\newblock \bibinfo{title}{From centralized to decentralized federated learning: Theoretical insights, privacy preservation, and robustness challenges}.
\newblock \bibinfo{journal}{arXiv preprint arXiv:2503.07505} .
\bibitem[{Li et~al.(2025b)Li, Li, Yu, Han and Jin}]{li2025financial}
\bibinfo{author}{Li, X.}, \bibinfo{author}{Li, W.}, \bibinfo{author}{Yu, X.}, \bibinfo{author}{Han, Z.}, \bibinfo{author}{Jin, Q.}, \bibinfo{year}{2025}b.
\newblock \bibinfo{title}{Financial risk assessment of imbalanced data based on nonlinear causal time-series network}.
\newblock \bibinfo{journal}{Information Processing \& Management} \bibinfo{volume}{62}, \bibinfo{pages}{104025}.
\bibitem[{Li et~al.(2024)Li, Wu, Zhang, Sun, Li and Wang}]{li2024adafgl}
\bibinfo{author}{Li, X.}, \bibinfo{author}{Wu, Z.}, \bibinfo{author}{Zhang, W.}, \bibinfo{author}{Sun, H.}, \bibinfo{author}{Li, R.H.}, \bibinfo{author}{Wang, G.}, \bibinfo{year}{2024}.
\newblock \bibinfo{title}{Adafgl: A new paradigm for federated node classification with topology heterogeneity}, in: \bibinfo{booktitle}{2024 IEEE 40th International Conference on Data Engineering (ICDE)}, \bibinfo{organization}{IEEE}. pp. \bibinfo{pages}{2517--2530}.
\bibitem[{Li et~al.(2023)Li, Wu, Zhang, Zhu, Li and Wang}]{li2023fedgta}
\bibinfo{author}{Li, X.}, \bibinfo{author}{Wu, Z.}, \bibinfo{author}{Zhang, W.}, \bibinfo{author}{Zhu, Y.}, \bibinfo{author}{Li, R.H.}, \bibinfo{author}{Wang, G.}, \bibinfo{year}{2023}.
\newblock \bibinfo{title}{Fedgta: Topology-aware averaging for federated graph learning}.
\newblock \bibinfo{journal}{Proceedings of the VLDB Endowment} \bibinfo{volume}{17}, \bibinfo{pages}{41--50}.
\bibitem[{Lian et~al.(2017)Lian, Zhang, Zhang, Hsieh, Zhang and Liu}]{lian2017can}
\bibinfo{author}{Lian, X.}, \bibinfo{author}{Zhang, C.}, \bibinfo{author}{Zhang, H.}, \bibinfo{author}{Hsieh, C.J.}, \bibinfo{author}{Zhang, W.}, \bibinfo{author}{Liu, J.}, \bibinfo{year}{2017}.
\newblock \bibinfo{title}{Can decentralized algorithms outperform centralized algorithms? a case study for decentralized parallel stochastic gradient descent}.
\newblock \bibinfo{journal}{Advances in neural information processing systems} \bibinfo{volume}{30}.
\bibitem[{Liu et~al.(2024)Liu, Shi, Li, Wu, Wang and Shen}]{liu2024decentralized}
\bibinfo{author}{Liu, Y.}, \bibinfo{author}{Shi, Y.}, \bibinfo{author}{Li, Q.}, \bibinfo{author}{Wu, B.}, \bibinfo{author}{Wang, X.}, \bibinfo{author}{Shen, L.}, \bibinfo{year}{2024}.
\newblock \bibinfo{title}{Decentralized directed collaboration for personalized federated learning}, in: \bibinfo{booktitle}{Proceedings of the IEEE/CVF conference on computer vision and pattern recognition}, pp. \bibinfo{pages}{23168--23178}.
\bibitem[{Mao et~al.(2025)Mao, Li, Xue, Li, Cai and Noman}]{MAO2025102876}
\bibinfo{author}{Mao, G.}, \bibinfo{author}{Li, H.}, \bibinfo{author}{Xue, L.}, \bibinfo{author}{Li, Y.}, \bibinfo{author}{Cai, Z.}, \bibinfo{author}{Noman, K.}, \bibinfo{year}{2025}.
\newblock \bibinfo{title}{Fedpm-sgn: A federated graph network for aviation equipment fault diagnosis by multi-sensor fusion in decentralized and heterogeneous setting}.
\newblock \bibinfo{journal}{Information Fusion} \bibinfo{volume}{117}, \bibinfo{pages}{102876}.
\bibitem[{McMahan et~al.(2017)McMahan, Moore, Ramage, Hampson and y~Arcas}]{mcmahan2017communication}
\bibinfo{author}{McMahan, B.}, \bibinfo{author}{Moore, E.}, \bibinfo{author}{Ramage, D.}, \bibinfo{author}{Hampson, S.}, \bibinfo{author}{y~Arcas, B.A.}, \bibinfo{year}{2017}.
\newblock \bibinfo{title}{Communication-efficient learning of deep networks from decentralized data}, in: \bibinfo{booktitle}{Artificial intelligence and statistics}, \bibinfo{organization}{PMLR}. pp. \bibinfo{pages}{1273--1282}.
\bibitem[{Onoszko et~al.(2021)Onoszko, Karlsson, Mogren and Zec}]{onoszko2021decentralized}
\bibinfo{author}{Onoszko, N.}, \bibinfo{author}{Karlsson, G.}, \bibinfo{author}{Mogren, O.}, \bibinfo{author}{Zec, E.L.}, \bibinfo{year}{2021}.
\newblock \bibinfo{title}{Decentralized federated learning of deep neural networks on non-iid data}.
\newblock \bibinfo{journal}{arXiv preprint arXiv:2107.08517} .
\bibitem[{Pei et~al.(2020)Pei, Wei, Chang, Lei and Yang}]{pei2020geom}
\bibinfo{author}{Pei, H.}, \bibinfo{author}{Wei, B.}, \bibinfo{author}{Chang, K.C.C.}, \bibinfo{author}{Lei, Y.}, \bibinfo{author}{Yang, B.}, \bibinfo{year}{2020}.
\newblock \bibinfo{title}{Geom-gcn: Geometric graph convolutional networks}, in: \bibinfo{booktitle}{8th International Conference on Learning Representations, ICLR 2020}.
\bibitem[{Pei et~al.(2021)Pei, Mao, Liu, Chen, Xu, Qiang and Tech}]{pei2021decentralized}
\bibinfo{author}{Pei, Y.}, \bibinfo{author}{Mao, R.}, \bibinfo{author}{Liu, Y.}, \bibinfo{author}{Chen, C.}, \bibinfo{author}{Xu, S.}, \bibinfo{author}{Qiang, F.}, \bibinfo{author}{Tech, B.E.}, \bibinfo{year}{2021}.
\newblock \bibinfo{title}{Decentralized federated graph neural networks}, in: \bibinfo{booktitle}{International workshop on federated and transfer learning for data sparsity and confidentiality in conjunction with IJCAI}.
\bibitem[{Qu et~al.(2022)Qu, Dai, Zhuang, Chen, Dong, Wu and Guo}]{qu2022decentralized}
\bibinfo{author}{Qu, Y.}, \bibinfo{author}{Dai, H.}, \bibinfo{author}{Zhuang, Y.}, \bibinfo{author}{Chen, J.}, \bibinfo{author}{Dong, C.}, \bibinfo{author}{Wu, F.}, \bibinfo{author}{Guo, S.}, \bibinfo{year}{2022}.
\newblock \bibinfo{title}{Decentralized federated learning for uav networks: Architecture, challenges, and opportunities}.
\newblock \bibinfo{journal}{IEEE Network} \bibinfo{volume}{35}, \bibinfo{pages}{156--162}.
\bibitem[{Roy et~al.(2019)Roy, Siddiqui, P{\"o}lsterl, Navab and Wachinger}]{roy2019braintorrent}
\bibinfo{author}{Roy, A.G.}, \bibinfo{author}{Siddiqui, S.}, \bibinfo{author}{P{\"o}lsterl, S.}, \bibinfo{author}{Navab, N.}, \bibinfo{author}{Wachinger, C.}, \bibinfo{year}{2019}.
\newblock \bibinfo{title}{Braintorrent: A peer-to-peer environment for decentralized federated learning}.
\newblock \bibinfo{journal}{arXiv preprint arXiv:1905.06731} .
\bibitem[{Shchur et~al.(2018)Shchur, Mumme, Bojchevski and G{\"u}nnemann}]{shchur2018pitfalls}
\bibinfo{author}{Shchur, O.}, \bibinfo{author}{Mumme, M.}, \bibinfo{author}{Bojchevski, A.}, \bibinfo{author}{G{\"u}nnemann, S.}, \bibinfo{year}{2018}.
\newblock \bibinfo{title}{Pitfalls of graph neural network evaluation}.
\newblock \bibinfo{journal}{arXiv preprint arXiv:1811.05868} .
\bibitem[{Sun et~al.(2022)Sun, Li and Wang}]{sun2022decentralized}
\bibinfo{author}{Sun, T.}, \bibinfo{author}{Li, D.}, \bibinfo{author}{Wang, B.}, \bibinfo{year}{2022}.
\newblock \bibinfo{title}{Decentralized federated averaging}.
\newblock \bibinfo{journal}{IEEE Transactions on Pattern Analysis and Machine Intelligence} \bibinfo{volume}{45}, \bibinfo{pages}{4289--4301}.
\bibitem[{Wu et~al.(2021)Wu, Wu, Cao, Huang and Xie}]{wu2021fedgnn}
\bibinfo{author}{Wu, C.}, \bibinfo{author}{Wu, F.}, \bibinfo{author}{Cao, Y.}, \bibinfo{author}{Huang, Y.}, \bibinfo{author}{Xie, X.}, \bibinfo{year}{2021}.
\newblock \bibinfo{title}{Fedgnn: Federated graph neural network for privacy-preserving recommendation}.
\newblock \bibinfo{journal}{arXiv preprint arXiv:2102.04925} .
\bibitem[{Wu et~al.(2022)Wu, Sun, Zhang, Xie and Cui}]{wu2022graph}
\bibinfo{author}{Wu, S.}, \bibinfo{author}{Sun, F.}, \bibinfo{author}{Zhang, W.}, \bibinfo{author}{Xie, X.}, \bibinfo{author}{Cui, B.}, \bibinfo{year}{2022}.
\newblock \bibinfo{title}{Graph neural networks in recommender systems: a survey}.
\newblock \bibinfo{journal}{ACM Computing Surveys} \bibinfo{volume}{55}, \bibinfo{pages}{1--37}.
\bibitem[{Xie et~al.(2021)Xie, Ma, Xiong and Yang}]{xie2021federated}
\bibinfo{author}{Xie, H.}, \bibinfo{author}{Ma, J.}, \bibinfo{author}{Xiong, L.}, \bibinfo{author}{Yang, C.}, \bibinfo{year}{2021}.
\newblock \bibinfo{title}{Federated graph classification over non-iid graphs}.
\newblock \bibinfo{journal}{Advances in neural information processing systems} \bibinfo{volume}{34}, \bibinfo{pages}{18839--18852}.
\bibitem[{Yang et~al.(2016)Yang, Cohen and Salakhudinov}]{yang2016revisiting}
\bibinfo{author}{Yang, Z.}, \bibinfo{author}{Cohen, W.}, \bibinfo{author}{Salakhudinov, R.}, \bibinfo{year}{2016}.
\newblock \bibinfo{title}{Revisiting semi-supervised learning with graph embeddings}, in: \bibinfo{booktitle}{International conference on machine learning}, \bibinfo{organization}{PMLR}. pp. \bibinfo{pages}{40--48}.
\bibitem[{Zhang et~al.(2021)Zhang, Yang, Li, Sun and Yiu}]{zhang2021subgraph}
\bibinfo{author}{Zhang, K.}, \bibinfo{author}{Yang, C.}, \bibinfo{author}{Li, X.}, \bibinfo{author}{Sun, L.}, \bibinfo{author}{Yiu, S.M.}, \bibinfo{year}{2021}.
\newblock \bibinfo{title}{Subgraph federated learning with missing neighbor generation}.
\newblock \bibinfo{journal}{Advances in neural information processing systems} \bibinfo{volume}{34}, \bibinfo{pages}{6671--6682}.
\bibitem[{Zhu et~al.(2022)Zhu, He, Zhang, Niu, Song and Tao}]{zhu2022topology}
\bibinfo{author}{Zhu, T.}, \bibinfo{author}{He, F.}, \bibinfo{author}{Zhang, L.}, \bibinfo{author}{Niu, Z.}, \bibinfo{author}{Song, M.}, \bibinfo{author}{Tao, D.}, \bibinfo{year}{2022}.
\newblock \bibinfo{title}{Topology-aware generalization of decentralized sgd}, in: \bibinfo{booktitle}{International Conference on Machine Learning}, \bibinfo{organization}{PMLR}. pp. \bibinfo{pages}{27479--27503}.
\bibitem[{Zhu et~al.(2024)Zhu, Li, Wu, Wu, Hu and Li}]{zhu2024fedtad}
\bibinfo{author}{Zhu, Y.}, \bibinfo{author}{Li, X.}, \bibinfo{author}{Wu, Z.}, \bibinfo{author}{Wu, D.}, \bibinfo{author}{Hu, M.}, \bibinfo{author}{Li, R.H.}, \bibinfo{year}{2024}.
\newblock \bibinfo{title}{Fedtad: Topology-aware data-free knowledge distillation for subgraph federated learning}.
\newblock \bibinfo{journal}{arXiv preprint arXiv:2404.14061} .

\end{thebibliography}

\end{document}